\pdfoutput=1

\documentclass[11pt]{article}

\usepackage[final]{acl}

\usepackage{times}
\usepackage{latexsym}

\usepackage[T1]{fontenc}

\usepackage[utf8]{inputenc}
\usepackage{url}            %
\usepackage{booktabs}       %
\usepackage{amsfonts}       %
\usepackage{nicefrac}       %
\usepackage{microtype}      %

\usepackage{adjustbox}
\usepackage{lipsum}
\usepackage{caption}
\usepackage{subcaption}
\usepackage{wrapfig}
\usepackage{array}
\usepackage[frozencache,cachedir=.]{minted}

\usepackage{titlesec}
\usepackage{lipsum}%

\usepackage{tikz}
\usetikzlibrary{automata,positioning}

\usepackage{inconsolata}

\usepackage{graphicx}

\title{Representing Rule-based Chatbots with Transformers}

\author{Dan Friedman\hspace{-4em} \And Abhishek Panigrahi \\Princeton Language and Intelligence, Princeton University\\
\texttt{\{dfriedman, ap34, danqic\}@cs.princeton.edu}\And \hspace{-4em}Danqi Chen}

\newcommand\ti[1]{\textit{#1}}

\newcommand\tf[1]{\textbf{#1}}
\newcommand\ttt[1]{\texttt{#1}}

\usepackage{amsmath,amsfonts}

\makeatletter
\newcommand{\ve}{\@ifnextchar\bgroup{\velong}{{\bm{e}}}}
\newcommand{\velong}[1]{{\bm{#1}}}
\makeatother

\def\ve{{\mathbf{e}}}

\def\vp{{\mathbf{p}}}

\DeclareMathAlphabet{\mathsfit}{\encodingdefault}{\sfdefault}{m}{sl}
\SetMathAlphabet{\mathsfit}{bold}{\encodingdefault}{\sfdefault}{bx}{n}

\def\tta{{\texttt{a}}}
\def\ttb{{\texttt{b}}}
\def\ttc{{\texttt{c}}}
\def\ttd{{\texttt{d}}}
\def\tte{{\texttt{e}}}
\def\ttf{{\texttt{f}}}
\def\ttg{{\texttt{g}}}
\def\tth{{\texttt{h}}}

\def\gV{{\mathcal{V}}}
\def\gW{{\mathcal{W}}}

\begin{document}
\maketitle

\begin{abstract}
What kind of internal mechanisms might Transformers use to conduct fluid, natural-sounding conversations? 
Prior work has illustrated by construction how Transformers can solve various synthetic tasks, such as sorting a list or recognizing formal languages, but it remains unclear how to extend this approach to a conversational setting.
In this work, we propose using ELIZA, a classic rule-based chatbot, as a setting for formal, mechanistic analysis of Transformer-based chatbots.
ELIZA allows us to formally model key aspects of conversation, including local pattern matching and long-term dialogue state tracking.
We first present a theoretical construction of a Transformer that implements the ELIZA chatbot.
Building on prior constructions, particularly those for simulating finite-state automata, we show how simpler mechanisms can be composed and extended to produce more sophisticated behavior.
Next, we conduct a set of empirical analyses of Transformers trained on synthetically generated ELIZA conversations.
Our analysis illustrates the kinds of mechanisms these models tend to prefer---for example, models favor an induction head mechanism over a more precise, position-based copying mechanism; and using intermediate generations to simulate recurrent data structures, akin to an implicit scratchpad or Chain-of-Thought.
Overall, by drawing an explicit connection between neural chatbots and interpretable, symbolic mechanisms, our results provide a new framework for the mechanistic analysis of conversational agents.\footnote{
Code and data are available at \url{https://github.com/princeton-nlp/ELIZA-Transformer}.
}

\end{abstract}

\section{Introduction}
\label{sec:introduction}

One approach to understanding 
Transformers~\cite{vaswani2017attention} is to identify explicit mechanisms that a Transformer could theoretically use to solve a particular task. 
This bottom-up approach has been used to characterize the expressivity of the Transformer architecture for a variety of synthetic and formal language tasks, including regular languages~\citep{bhattamishra2020ability,liu2023transformers}, Dyck languages~\citep{yao2021self}, and PCFGs~\citep{zhao2023transformers}. 
However, this line of work has focused mainly on simple algorithmic tasks applied to single-sentence inputs, and it remains an open question whether we can extend these approaches to understand how Transformers could conduct natural-sounding conversations.
In this work, we propose to use \ti{rule-based chatbots} for formal and mechanistic analysis of neural conversational agents.
We first present theoretical constructions of how a Transformer can implement a classic rule-based chatbot algorithm, and then we use these constructions to guide a series of empirical investigations into how Transformers learn to solve such tasks when they are trained on synthetic conversation data.

\begin{figure*}[t]
\centering
\includegraphics[width=\linewidth]{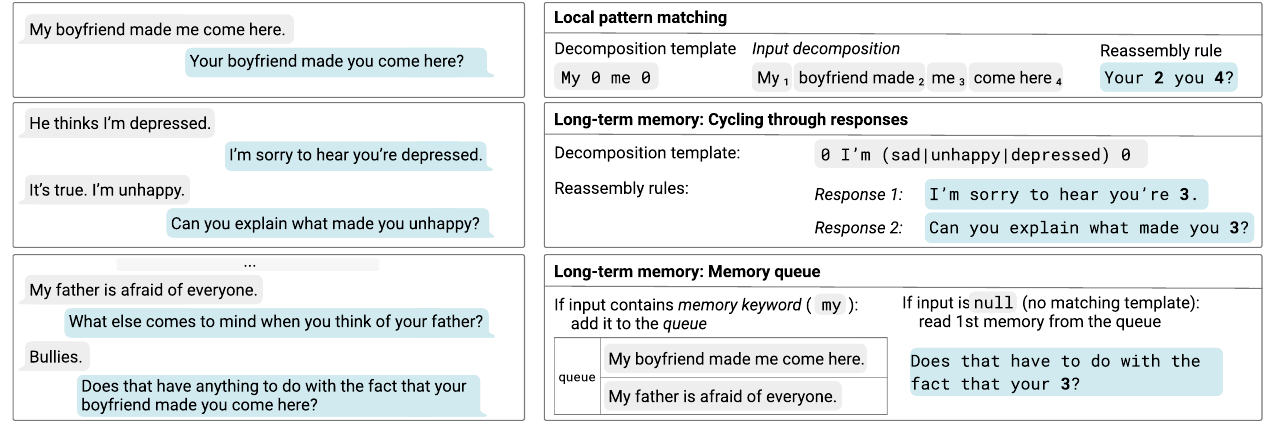}
\caption{
An example ELIZA conversation, adapted from~\citet{weizenbaum1966eliza} (\textit{left}) and the corresponding parts of the ELIZA program (\textit{right}).
ELIZA uses both local pattern matching and two long-term memory mechanisms (\ti{cycling through responses}, and a \emph{memory queue}).
At each turn, ELIZA compares the most recent input to a set of \ti{decomposition templates} and applies one of the associated \ti{reassembly rules}.
The \ttt{0} symbols in the decomposition template are wildcards, which are used to decompose the user's input into segments.
A response is generated by replacing each numeral in the reassembly rule with corresponding segment of the user's input.
If a template is matched more than once in a conversation, ELIZA cycles through a list of possible reassembly rules. %
If the input contains a special keyword (``my''), ELIZA stores it in a {memory queue};
later, if an input does not match any of the templates, ELIZA reads the first memory from the queue.
}
\label{fig:eliza_example}
\end{figure*}

In particular, we focus on ELIZA~\citep{weizenbaum1966eliza}, one of the first artificial chatbots.
The ELIZA algorithm is simple but exhibits a number of sophisticated conversational behaviors (Fig.~\ref{fig:eliza_example}).
The majority of ELIZA's behavior is based on local pattern/transformation rules: ELIZA compares the user's input to a set of templates, and responds by reassembling the input according to an associated transformation rule.
However, ELIZA also employs several mechanisms that make use of the full conversational history, including a mechanism for varying its responses between successive turns, and a \emph{memory queue} to refer to turns from the beginning of the conversation.
ELIZA therefore offers a natural next step from simpler, sentence-level settings, comprising both local pattern matching and long-distance dialogue state tracking.

First, we describe how a decoder-only Transformer could implement the ELIZA program (Fig.~\ref{fig:overview}).
We start by showing how we can use constructions from prior work as modular building blocks---in particular, by decomposing the task into a cascade of finite state automata~\citep{liu2023transformers,yang2024masked}, along with a copying mechanism for generating responses.
This decomposition attests to the usefulness of algebraic automata as building blocks for characterizing complex behavior in Transformers.
On the other hand, we also identify alternative constructions for key subtasks, including a more robust copying mechanism (Sec.~\ref{sec:cycling}) and memory mechanisms (Sec.~\ref{sec:memory_queue}) that make use of intermediate ELIZA outputs---akin to a scratchpad~\citep{nye2021show} or Chain-of-Thought~\citep{wei2022chain}.
These alternative constructions inform our empirical investigations later on.
Incidentally, the ELIZA framework happens to be Turing complete~\citep{hay2022eliza}; our results therefore lead to a simple, alternative construction for a Transformer that simulates a Turing machine, which we discuss in Appendix~\ref{app:turing_machine}.

Next, we generate a dataset of ELIZA transcripts and train Transformers to simulate the ELIZA algorithm (Sec.~\ref{sec:experiments}).
We investigate which aspects of the task are difficult for the models to learn, and find that models struggle the most with precise copying and with the memory queue mechanism---which requires the composition of several distinct mechanisms. %
We further study which of our hypothesized constructions better match what the models learn, and how the result varies according to the data distribution. %
For copying, we find that models have a strong bias for an induction head mechanism~\citep{olsson2022context}, leading to worse performance on sequences with a high degree of internal repetition.
For the memory components, we find that models make use of intermediate outputs to simulate the relevant data structures, which underscores the importance of considering intermediate computation in understanding Transformers, even without an explicit scratchpad or Chain-of-Thought.
Together, our results illustrate that ELIZA offers a rich setting for mechanistic analysis of learning dynamics, allowing us to decompose the task into subtasks, conduct fine-grained behavioral analysis, and connect this analysis to predictions about the model's mechanisms.

By drawing an explicit connection between neural chatbots and interpretable, symbolic mechanisms, our results offer a new setting for an algorithm-level understanding of conversational agents.
We conclude by discussing the broader implications of our results for future work on interpretability and the science of language models.

\section{Background: ELIZA}
\label{sec:background}
We start by describing the ELIZA algorithm~\citep{weizenbaum1966eliza}, following the presentation of~\citep{jurafsky2020chatbots}.
The ELIZA algorithm can be decomposed into two types of behavior: local pattern matching and long-term memory, illustrated in Fig.~\ref{fig:eliza_example}.
We discuss ELIZA in more detail in Appendix~\ref{app:eliza_details}.

\subsection{Local Pattern Matching}
First, ELIZA compares the most recent user input to an inventory of pattern/transformation rules, such as the following:
\begin{align*}
    \texttt{0 YOU 0 ME} \to \text{What makes you think I \ttt{3} you?}
\end{align*}
The left-hand side of the rule is called a \ti{decomposition template} and corresponds to a simple regular expression, where the $\texttt{0}$ symbol is a wildcard that matches 0 or more occurrences of any word.
If an input matches a template, it is partitioned into a set of \ti{decomposition groups} corresponding to the wildcards.
For example, the input {``It seems like you hate me''} would be decomposed into four groups: (1) {It seems like} (2) {you} (3) {hate} (4) {me}.
The right-hand side of the rule is called a \ti{reassembly rule}, and a response is generated by replacing any number in the reassembly rule with the content of the corresponding decomposition group.
In this case, ELIZA will respond, ``What makes you think I hate you?''
An ELIZA chatbot is defined by an inventory of these rules, which are organized into a configuration file known as the \ti{script}.
Each decomposition template is assigned a rank and associated with one or more reassembly rules.
Given an input, ELIZA finds the highest ranked template that matches the sentence and applies one of the associated reassembly rules.
The script also must assign some reassembly rules to a \ti{null} template, which is used if none of the templates matches.

\subsection{Long-Term Memory}
While most responses consider only the previous utterance, ELIZA includes two mechanisms that use information from earlier in the conversation.

\paragraph{Cycling through reassembly rules}
First, each template in a script can be associated with a list of reassembly rules.
If the template is matched multiple times in a conversation, ELIZA will cycle through all the rules in the list before returning to the first item.
For example, in Weizenbaum's ELIZA script, if the input contains the word ``sorry,'' ELIZA will initially respond with ``Please don't apologize.''
If the user says ``sorry'' a second time, ELIZA will say ``Apologies aren't necessary.''
If the user continues with ``sorry'', ELIZA will eventually say ``I've told you that apologies are not required,'' and then cycle back to the first rule in the list.

\paragraph{Memory queue}
Second, if an utterance contains a particular keyword (by default, the word ``my''), ELIZA stores it in a queue, referred to as the \ti{memory queue}.
Later in the conversation, if the user's input does not match any of the templates, ELIZA will output the first item in the queue, applying one of a set of memory reassembly rules.
For example, at the beginning of the conversation in Fig.~\ref{fig:eliza_example}, the user states ``My boyfriend made me come here.''
Many turns later, the user enters a sentence that does not match any of the patterns, and ELIZA replies, ``Does that have anything to do with the fact that your boyfriend made you come here?''

\begin{figure*}[t]
\centering
\includegraphics[width=0.9\linewidth]{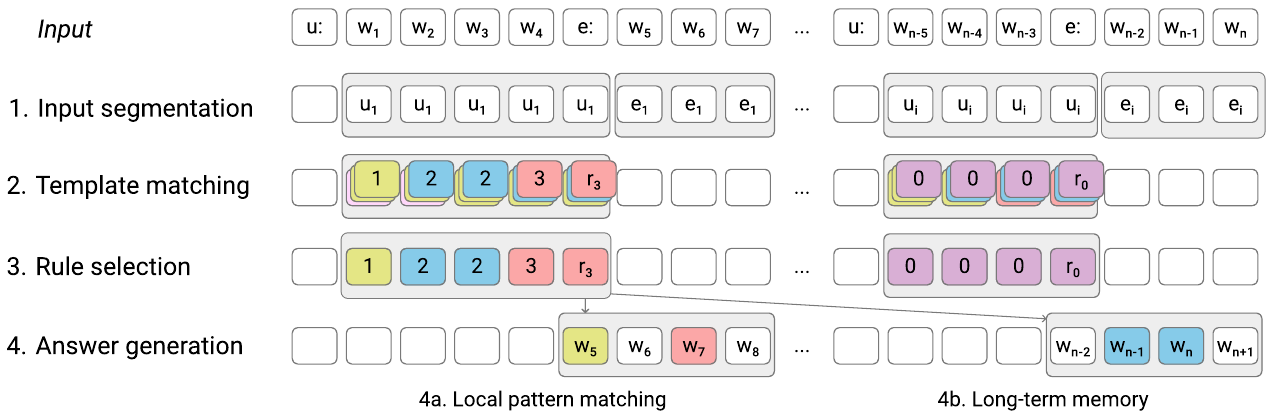}
\caption{
The input to the Transformer is the conversation history, consisting of user inputs (beginning with \ttt{u:}) followed by ELIZA's responses (\ttt{e:}).
The constructions then have four parts.
First, the input is divided into segments, each corresponding to a user input or ELIZA response.
Second, the model attempts to match each user input to a \ti{decomposition template}; this step is executed in parallel, with each input compared to every possible decomposition template.
The model then identifies the highest scoring template and selects a \ti{reassembly rule}, taking into account the number of times this template has been matched earlier in the conversation.
Finally, the model generates an answer, either by applying a reassembly rule to the most recent user input (\ti{4a}) or by transforming an input from earlier in the conversation, using the ``memory queue'' mechanism (\ti{4b}).
}
\label{fig:overview}
\end{figure*}

\section{Constructions}
\label{sec:construction}

Now we present our constructions for implementing the ELIZA program with a Transformer decoder.
We build the construction modularly by composing constructions for each subtask in ELIZA (Fig.~\ref{fig:overview}).
For the key subtasks, we identify multiple possible mechanisms a Transformer could use, some building on prior constructions---in particular, for simulating finite state automata---and others drawing on intuitions from methods like Chain-of-Thought~\citep{wei2022chain}.
In Section~\ref{sec:experiments}, we will investigate empirically which of these options are a better match for what the models learn.
We describe the constructions at a high level in this section and defer the details to Appendix~\ref{app:construction_details}.

\paragraph{Setup}
We consider a decoder-only Transformer with softmax attention.
At each turn in the conversation, the input will be the concatenation of the conversation so far, with each user input and each ELIZA response preceded by a special delimiter character, either \ttt{u:} (user) or \ttt{e:} (ELIZA), respectively.
The constructions use no positional encodings, as we can use the self-attention mask to infer positional information~\citep{haviv2022transformer,kazemnejad2023impact}, and to segment the input into turns, in order to restrict attention to a particular utterance.
See Appendix~\ref{app:input_segmentation} for more details.

\subsection{Local Pattern Matching}
\label{sec:template_matching}

We start by considering a single turn in the conversation, which involves first finding a template that matches the input, and then generating a response using the associated transformation rule.

\paragraph{Matching templates}
For template matching, we make use of the fact that ELIZA templates are equivalent to star-free regular expressions~\citep{mcnaughton1971counter}; these can be recognized by simulating a corresponding finite-state automaton.
We build on the constructions of~\citep{liu2023transformers,yang2024masked}. 
At a high level, we can recognize a template with $L$ symbols using a Transformer with $L$ layers.
At each layer $\ell$ and position $i$, the Transformer determines whether the input matches the first $\ell$ symbols of the template at position $i$.
The final output can be used to both (a) determine if an input matches a template, and (b) decompose the input according to the template's decomposition groups.
Our constructions recognize multiple templates in parallel using two attention heads per layer---one attending uniformly to the full prefix, and one attending to the previous position.
The depth of the Transformer therefore scales with the length of the longest template in the configuration script, and the width scales with the total number of templates in the script.
See Appendix~\ref{app:template_matching} for more details.

\paragraph{Generating a response}
Now we assume that we have identified a matching template and that the embedding for each input token identifies the decomposition group to which that token belongs.
The next step is to apply the reassembly rule to the input to generate a response.
At each generation step, the model needs to either generate a constant word (defined by the reassembly rule), or copy a word from one of the decomposition groups. %
We present two high-level options, deferring the precise details to Appendix~\ref{app:generation}.

\paragraph{Option 1: Content-based attention (induction head)}
The first possible approach is based on the induction head~\citep{olsson2022context}.
This mechanism has been widely studied in prior work and is considered a key primitive in Transformers~\citep[e.g. ][]{reddy2024mechanistic,singh2024needs,akyurek2024context,edelman2024evolution}.
In our setting, we define an induction head as follows: Given an input sequence $w$, at each output position $i$, an induction head attends to an input position $j$ such that $w_{i-n}, \ldots, w_i = w_{j-n-1}, \ldots, w_{j-1}$, and copies the token value $w_j$ (where $n$ is some context size).
This mechanism has a drawback: as noted by~\citet{zhou2023algorithms}, it assumes that each word has a unique $n$-gram prefix, so it can fail if the same $n$-gram appears more than once in the input sequence. Appendix Table  \ref{tab:induction_head_reassembly} shows an example.

\paragraph{Option 2: Position-based attention}
To avoid this shortcoming, we propose a second option that uses position rather than content to identify the next word to copy.
At each step, we can identify the position to copy next as a function of the reassembly rule; the number of tokens generated so far; and the number of tokens in each decomposition group.
This can be accomplished using an attention layer to obtain the relevant counts, and a feedforward layer to calculate the target position
(See Appendix Fig.~\ref{fig:generate_token} for details).
Compared to the induction head, this mechanism works equally well regardless of the content of the copying segment.
The drawback is that it relies on precise position arithmetic;~\citet{zhou2023algorithms} argue that such mechanisms are difficult for Transformers to learn, and might not generalize to longer sequences.

\subsection{Cycling through Reassembly Rules}
\label{sec:cycling}
Now we turn to the first subtask that makes use of information from earlier in the conversation: \emph{cycling through reassembly rules}.
Specifically, we allow each template $t$ to be associated with a sequence of reassembly rules $r_1, \ldots, r_M$.
When template $t$ appears in a conversation for the $i^{th}$ time, the model should respond with rule $r_{i \% M}$.
We consider two mechanisms (see Appendix Fig.~\ref{fig:alternatives} for an illustration).

\paragraph{Option 1: Modular prefix sum} 
One option is to use the modular prefix sum mechanism described by~\citet{liu2023transformers}: an attention head counts the number of times $t$ has been matched, and an MLP outputs the result modulo $M$.
We anticipate that such a mechanism might perform worse as the sequence grows longer, as the model must attend over a longer sequence and process a larger count.
Additionally, different templates can have a different numbers of reassembly rules, so the model must learn a separate modulus for each template.

\paragraph{Option 2: Intermediate outputs}
The model can avoid modular arithmetic by making use of its earlier outputs.
Specifically, the model can reuse the template matching mechanism to identify outputs where it responded to template $t$ with any of $r_1, \ldots, r_M$.
The model can then attend to the most recent of these responses $r_i$, and respond with $r_{(i+1) \% M}$.
This mechanism works regardless of the cycle number.
However, it would fail if the same reassembly rule appears more than once in the list, or if the reassembly rules are difficult to identify.
This use of intermediate outputs resembles prompting methods like scratchpad~\citep{nye2021show} and Chain-of-Thought~\citep{wei2022chain}, which we discuss in more detail in Sec.~\ref{sec:experiments}.

\subsection{Memory Queue}
\label{sec:memory_queue}
Finally, we incorporate the memory queue component.
Recall that ELIZA adds a user input to the memory queue if it contains a special memory keyword (e.g. ``my'') and matches an associated template.
ELIZA reads an item from the memory queue if (a) the most recent input does not match any templates and (b) the queue is not empty.
Given the output of the template-matching stage, it is simple to determine whether an input represents an \ttt{enqueue} event or a \ttt{no\_match} event.
The main challenge is to determine whether there are any items in the queue, and so whether a given \ttt{no\_match} input should trigger a \ttt{dequeue}.
Again, we present two mechanisms, illustrated in Fig.~\ref{fig:alternatives}.

\paragraph{Option 1: Gridworld automaton}
The first approach we consider is to use the construction from~\citet{liu2023transformers} for simulating a one-dimensional ``gridworld'' automaton, which has $S$ numbered states and two actions: ``increment the state if possible'' and ``decrement the state if possible.''
At each \ttt{enqueue} event, the automaton increments the state if possible, and at each \ttt{no\_match} event, the model decrements the state if possible.
If the state is decremented, we can conclude that this input should trigger a \ttt{dequeue}.
We can then calculate the number of dequeues in the sequence, $d$, and read the $d^{th}$ memory in the queue.
~\citet{liu2023transformers} present a gridworld construction with two Transformer layers and $2S$ attention heads, which would allow us to implement a memory queue with a maximum size of $S$.

\paragraph{Option 2: Intermediate outputs}
Alternatively, as above, we can instead identify \ttt{dequeue} operations by examining earlier ELIZA outputs.
By reusing the template matching mechanism, we can check whether an ELIZA response matches one of reassembly rules associated with the dequeue operation.
Then, letting $d$ denote the number of dequeue operations, if $d$ is less than the number of enqueue operations, we read the $d^{th}$ memory from the queue.
Compared to the gridworld approach, this construction  uses fewer attention heads and does not limit the size of the memory queue, but it does impose a limit on the total number of enqueues (because we need to embed the number of enqueues to attend to the right memory). %

\section{Experiments}
\label{sec:experiments}

Now we investigate how Transformers learn the ELIZA program in practice when we train them on conversation transcripts.
We focus on investigating how the model solves the subtasks for which we identified more than one possible construction.

\begin{figure*}[t]
\centering
\begin{subfigure}[b]{0.55\textwidth}
    \centering
   \includegraphics[width=\textwidth]{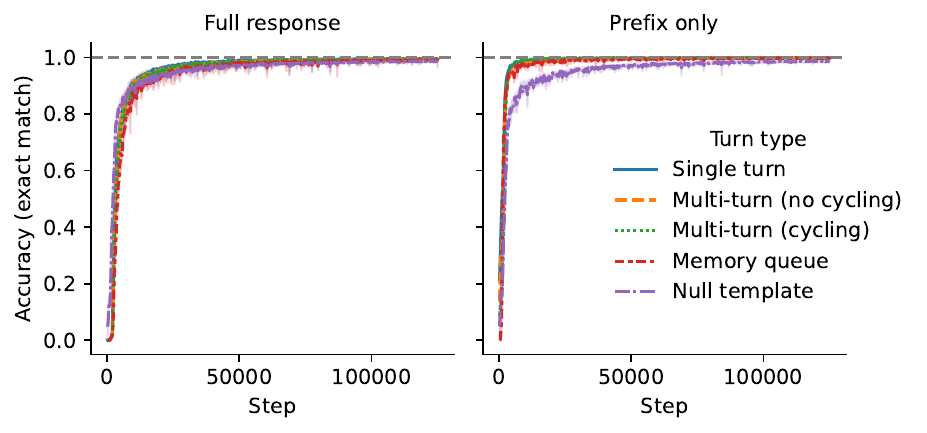}
    \caption{Accuracy (training curve).}
    \label{fig:accuracy_by_turn_type_line}
\end{subfigure}
\hfill
\begin{subfigure}[b]{0.42\textwidth}
    \centering
    \includegraphics[width=\textwidth]{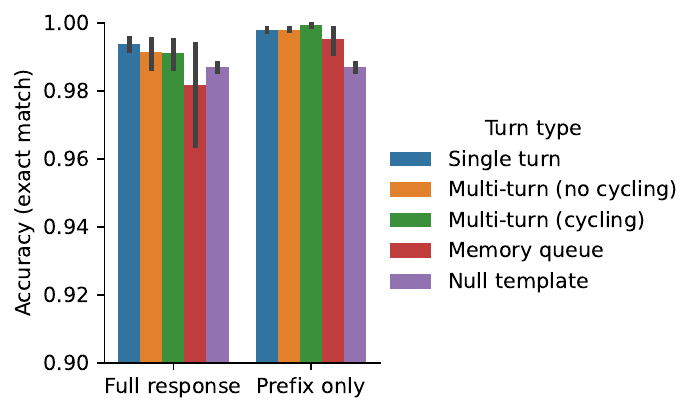}
    \caption{Accuracy (end of training).}
    \label{fig:accuracy_by_turn_type_bar}
\end{subfigure}
\caption{
Turn-level accuracy of Transformers trained on ELIZA conversations over training (Fig.~\ref{fig:accuracy_by_turn_type_line}) and at the final checkpoint (Fig.~\ref{fig:accuracy_by_turn_type_bar}), for models trained with three random seeds.
Transformers quickly learn to identify the correct reassembly rule (measured by \ti{Prefix only} accuracy), and take longer to learn to implement the transformation correctly (\ti{Full response}).
Accuracy is slightly worse on multi-turn and memory queue examples; see \S\ref{app:learning}.
}
\label{fig:accuracy_by_turn_type}
\end{figure*}

\label{sec:experiment_setup}
\paragraph{Generating data}
For these experiments, we generate synthetic ELIZA data.
We first sample a configuration script consisting of templates and associated reassembly rules, and then generate conversations that exhibit the different rules.
At each turn, we sample a template, and then sample a sentence that matches that template by replacing each wildcard with a sequence of words sampled uniformly from the vocabulary, and then generating a response according to the ELIZA rules.
The vocabulary consists of the 26 lowercase letters.
More details are provided in Appendix~\ref{app:data_generation_details}.

\paragraph{Model and training}
We train Transformers with 8 layers, 12 attention heads per layer, and a hidden size of 768.
We use the GPT-2 architecture~\cite{radford2019language} but remove the position encodings and train all models from scratch.
The models are trained to predict the ELIZA responses (and not the user inputs).
See Appendix~\ref{app:models_and_training} for more details.

\subsection{Which Parts of the ELIZA Program are Harder to Learn?}
\label{app:learning}
In Figure~\ref{fig:accuracy_by_turn_type}, we plot the accuracy over the course of training and at the final checkpoint.
The \ti{Full response} accuracy is the per-turn exact match accuracy.
The \ti{Prefix only} accuracy is the accuracy on the two-word prefix of the response, which we ensure is unique for each reassembly rule.
This metric provides a proxy for distinguishing whether errors are due to either (a) failure to identify the correct rule, or (b) failure to implement the rule correctly.
We additionally break down the results by turn type, defined as follows:
\ti{Single-turn:} The first response in the conversation.
\ti{Multi-turn (no cycling):} The response for the first instance of a template in the conversation.
\ti{Multi-turn (cycling):} The response for a template that has already appeared at least once in the conversation.
\ti{Memory queue:} Responses that read from the memory queue.
\ti{Null template:} Responses to inputs that do not match any templates, when the memory queue is empty.

\paragraph{Accuracy by subtask}
Figure~\ref{fig:accuracy_by_turn_type_line} shows that the models quickly learn to identify the correct action (as measured by prefix accuracy), achieving near-perfect accuracy on almost all categories.
The exception is the null template, which is used when the input does not match any other pattern and the memory queue is empty.
At the final checkpoint (Fig.~\ref{fig:accuracy_by_turn_type_bar}), accuracy is high but still imperfect, with slightly lower accuracy in the multi-turn setting.

\begin{figure*}[t]
\centering
\begin{subfigure}[b]{0.24\textwidth}
    \centering
    \includegraphics[width=\textwidth]{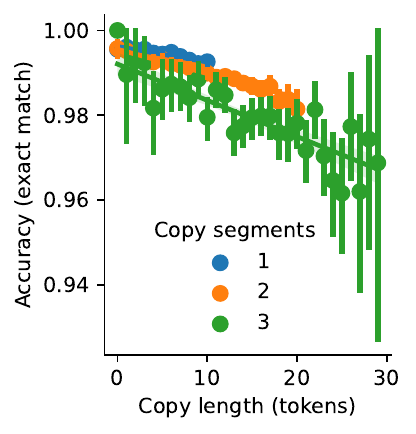}
    \caption{Copying.}
    \label{fig:accuracy_by_copy_length}
\end{subfigure}
\hfill
\begin{subfigure}[b]{0.48\textwidth}
    \centering
    \includegraphics[width=\textwidth]{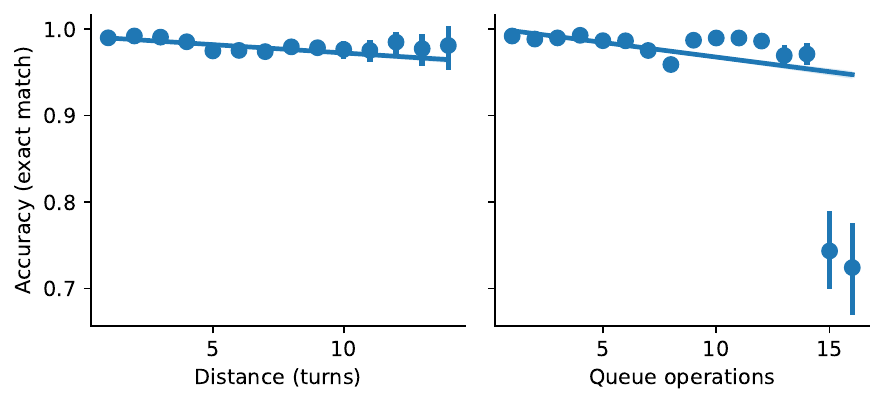}
    \caption{Dequeues.}
    \label{fig:accuracy_memory_queue}
\end{subfigure}
\begin{subfigure}[b]{0.24\textwidth}
    \centering
    \includegraphics[width=\textwidth]{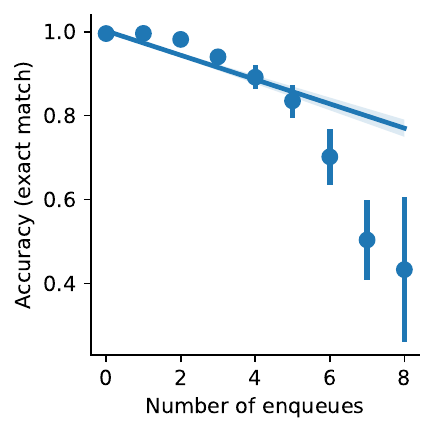}
    \caption{\ti{Null} inputs.}
    \label{fig:accuracy_null_template}
\end{subfigure}
\caption{
Which aspects of the task are most difficult for Transformers to learn?
\tf{Copying} (Fig.~\ref{fig:accuracy_by_copy_length}): Accuracy decreases considerably with the number of tokens to copy, and decreases slightly with the number of distinct copying segments. 
\tf{Memory queue} (Fig.~\ref{fig:accuracy_memory_queue}): The dequeue accuracy decreases when there is a greater distance to the target memory and when there have been more queue operations earlier in the sequence.
\tf{Null template} (Fig.~\ref{fig:accuracy_null_template}): The models do perfectly on \ti{null} inputs provided there have been no memory turns in the sequence; accuracy decreases with the number of enqueues, indicating that the models struggle when the queue has been used but is now empty.
}
\label{fig:error_analysis}
\end{figure*}

\paragraph{Error analysis}
In Figure~\ref{fig:error_analysis}, we test whether the model's errors are correlated with various properties of the input.
We identify two main issues.
First, the models seem to struggle with precise copying.
Fig.~\ref{fig:accuracy_by_copy_length} shows that accuracy is strongly correlated with the number of tokens the model has to copy, and only slightly correlated with the complexity of the decomposition rule (defined as the number of distinct copying segments in the transformation).
Similarly, Fig.~\ref{fig:accuracy_memory_queue} (\ti{left}) shows that memory queue accuracy decreases with the distance between the current turn and the target memory, perhaps indicating issues with long-distance copying.

Second, some errors seem to be related to tracking the state of the memory queue.
Fig.~\ref{fig:accuracy_memory_queue} (\ti{right}) shows that accuracy is negatively correlated with the total number of enqueue and dequeue operations in the sequence.
Fig.~\ref{fig:accuracy_null_template} shows that the model performs perfectly on null inputs, provided that there have been no memory turns; accuracy decreases with the number of enqueues, indicating that the models struggle when the queue has been used but is now empty.
 We investigate this result in more detail in Appendix Sec.~\ref{app:null_inputs}.

\begin{figure*}[t]
\centering
\begin{subfigure}[b]{0.22\textwidth}
    \centering
    \includegraphics[width=\textwidth]{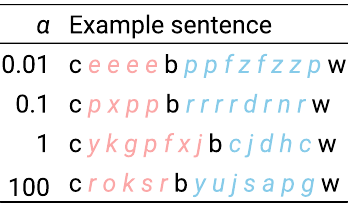}
    \caption{Examples.}
    \label{fig:unigram_concentration_examples}
\end{subfigure}
\hfill
\begin{subfigure}[b]{0.32\textwidth}
    \centering
    \includegraphics[width=\textwidth]{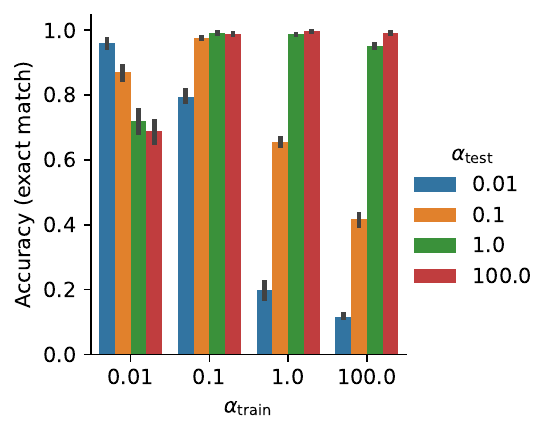}
    \caption{Comparing generalization.}
    \label{fig:unigram_concentration_accuracy}
\end{subfigure}
\hfill
\begin{subfigure}[b]{0.4\textwidth}
    \centering
    \includegraphics[width=\textwidth]{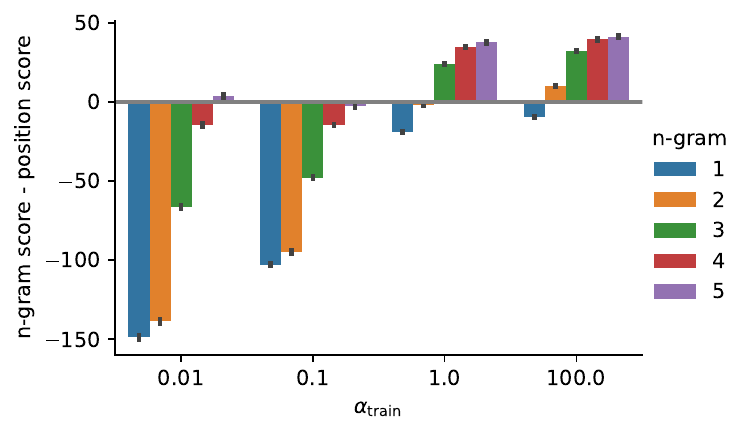}
    \caption{Comparing mechanisms.}
    \label{fig:unigram_concentration_attention_difference}
\end{subfigure}
\caption{
We train and test models on datasets that vary in whether copying segments are more or less likely to contain the same $n$-gram multiple times (Fig.~\ref{fig:unigram_concentration_examples}).
Models generalize poorly to data with more or less repetition compared to the training distribution (Fig.~\ref{fig:unigram_concentration_accuracy}).
Fig.~\ref{fig:unigram_concentration_attention_difference} suggests that models trained on less repetitive data assign higher attention scores to tokens with matching contexts, rather than calculating the correct target position. See \S\ref{sec:mechanisms}.
}
\label{fig:copying_mechanisms}
\end{figure*}

\subsection{Comparing Copying Mechanisms}
\label{sec:mechanisms}

In Section~\ref{sec:template_matching}, we identified two possible mechanisms for copying: an induction head, which attends based on the content of the input, and a position-based attention head.
We predicted that the induction head will fail when the same $n$-gram appears more than once in the input, while the position-based mechanism will generalize.
To explore which mechanism the models seem to learn, we generate (single-turn) datasets that vary in how likely it is for the same $n$-gram to appear multiple times in a sequence.
This is controlled by a parameter $\alpha$, with $\alpha < 1$ corresponding to more repetition of $n$-grams and $\alpha > 1$ making it more likely that most $n$-grams are unique.\footnote{
Specifically, given a template, we generate a sentence as follows:
For each wildcard in the sentence, we sample a vector $\vp \sim \mathrm{Dirichlet}(\alpha \mathbf{1})$, where $\mathbf{1}$ is a 26-dimensional vector of all 1's and $\alpha$ is the \ti{concentration} parameter.
Then we replace the wildcard with 0-20 words sampled from $\mathrm{Categorical}(\vp)$.
With $\alpha < 1$, $\vp$ is more likely to concentrate most probability on a small number of items, meaning each segment is more likely to contain repeated $n$-grams.
With $\alpha > 1$, $\vp$ is more likely to be close to the uniform distribution (corresponding to our setting in the previous section).
See Fig.~\ref{fig:unigram_concentration_examples} for examples.
}

We start by training models on four different datasets (Fig.~\ref{fig:unigram_concentration_examples}) and evaluating how well they generalize to datasets with more or less repetition (Fig.~\ref{fig:unigram_concentration_accuracy}).
The model trained with the least amount of repetition ($\alpha = 100$) performs well in-domain but suffers severe degradation on data with more repetition;
this provides preliminary evidence that, in our default setting, models learn an induction head mechanism that does not generalize when $n$-grams can repeat.
On the other hand, models trained on the most repetitive data ($\alpha = 0.01$) generalize poorly to higher values of $\alpha$.
The best-generalizing model is trained with an $\alpha = 0.1$, suggesting that some moderate amount of repetition is needed to learn a robust mechanism.
In Appendix Fig.~\ref{fig:unigram_concentration_accuracy_training_curves}, we plot results over the course of training, indicating that the most repetitive data also takes longer to learn.

\begin{figure*}[t]
\centering
\begin{subfigure}[b]{0.42\textwidth}
    \centering
    \includegraphics[width=\textwidth]{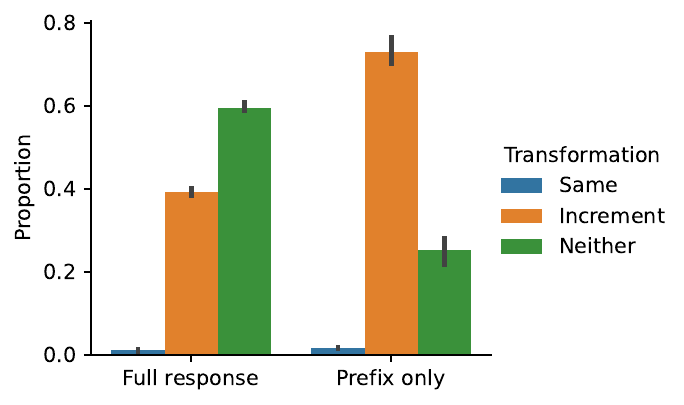}
    \caption{Cycling through reassembly rules.}
    \label{fig:cycling_counterfactual}
\end{subfigure}
\hfill
\begin{subfigure}[b]{0.42\textwidth}
    \centering
    \includegraphics[width=\textwidth]{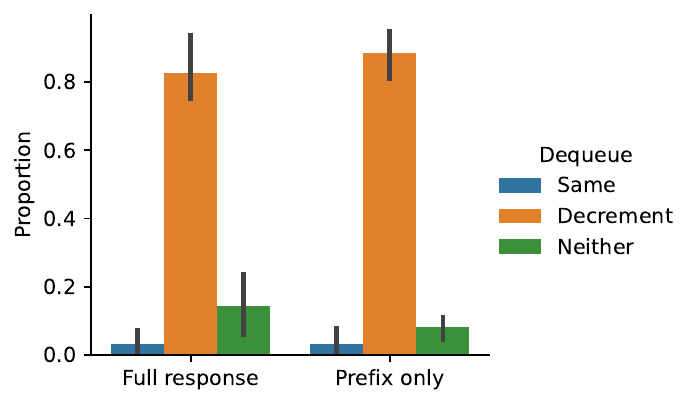}
    \caption{Memory queue.}
    \label{fig:memory_queue_counterfactual}
\end{subfigure}
\caption{
We design counter-factual experiments to test whether models make use of intermediate generations to keep track of the response cycle (Fig.~\ref{fig:cycling_counterfactual}) or memory queue (Fig.~\ref{fig:memory_queue_counterfactual}), or rely only on the user inputs.
Error bars show 95\% confidence interval over models trained with three random seeds.
Both experiments indicate that the models use their own outputs from earlier in the sequence.
When we edit the model's earlier output, we can reliably influence it to increment the response cycle or read a memory from earlier in the queue.
\vspace{-1em}
}
\label{fig:counterfactuals}
\end{figure*}

To get a sense of what mechanism these models actually learn, we examine the final layer attention heads.
Specifically, given an ELIZA response, for each output position $i$, we calculate the position $j$ of the input token that should be copied next.
Then we calculate the average pre-softmax attention score between the query embedding at position $i$ and key embeddings drawn from other validation examples that satisfy one of two conditions:
either the key has same $n$-gram prefix as the query $i$, but appears at a position $k \neq j$; or the key appears at the target position $j$ but has a different $n$-gram prefix ($w_{i-n:i} \neq w_{j-n-1:j-1}$).

In Figure~\ref{fig:unigram_concentration_attention_difference}, we plot the difference between these scores for different $n$-gram windows, averaging over attention heads, with positive values indicating that the model assigns higher scores to content than position.
(We plot the results for each head individually in Appendix Fig.~\ref{fig:unigram_concentration_attention_difference_by_head}.)
When $\alpha \geq 1$, the models prefer content to position once there is a prefix match of at least three tokens in length.
For all models, the content score increases with the length of the matching $n$-gram, with a steeper increase when $\alpha < 1$.
The model trained with a moderate amount of repetition ($\alpha = 0.1$) generalizes the best and is also the only model that prefers position to content even at the longest context window.
While all models are sensitive to content to some extent, the results illustrate how the data distribution influences which mechanism the model uses, and how well they generalize.

\subsection{Comparing Memory Mechanisms}
Next, we examine which mechanism the models learn for the two subtasks that rely on information from earlier in the conversation: cycling through reassembly rules, and the memory queue.
In Sections~\ref{sec:cycling} and~\ref{sec:memory_queue}, we offered two possible constructions for each subtask: one construction based on simulating an automaton and one based on processing previously generated outputs.
Here, we design counter-factual experiments to test whether the model is sensitive to earlier responses.
For each mechanism, we edit the model's response to an intermediate turn in the sequence and then test the model's response at a subsequent turn
(see Appendix~\ref{app:mechanism_analysis} for details).
In Fig.~\ref{fig:counterfactuals}, we test whether the response is consistent with the automaton construction, which predicts that the reponse will be unchanged (\ti{Same}); the intermediate-output construction, which predicts that the response will change in a specific way---either incrementing the cycle counter (\ti{Increment}) or reading a memory from earlier in the clue (\ti{Decrement}); or whether it matches neither prediction (\ti{Neither}).
In both cases, the model's behavior is most consistent with the intermediate-output hypothesis, either incrementing the cycle counter or decrementing the memory queue counter as predicted.
This illustrates the importance of considering intermediate outputs in understanding Transformer behavior, even without an explicit scratchpad or Chain-of-Thought.

\section{Discussion and Related Work}
\label{sec:related_work}

\paragraph{Expressivity with formal languages}
Numerous works have formalized the expressive power of Transformers on formal languages. \citet{perez2021attention,perez2019turing,bhattamishra2020computational} show that Transformers with hard attention are Turing complete, and~\citet{wei2022statistically} study their statistical learnability. \citet{merrill2022saturated,merrill2023logic,hao2022formal,hahn2020theoretical} further distinguish the expressivity of transformers with different hard attention patterns.
Other works have investigated encoding specific algorithms in smaller simulators, e.g. bounded-depth Dyck languages \citep{yao2021self}, modular prefix sums \citep{anil2022exploring}, adders \citep{nanda2023progress}, regular languages \citep{bhattamishra2020ability}, sparse logical predicates \citep{edelman2022inductive}, and $n$-gram language models~\citep{svete2024transformers}. \citet{liu2023transformers} propose a unified theory for expressivity of different automata with transformers. 
We refer the readers to~\citet{strobl2024formal} for a more comprehensive survey.

Building on these works, numerous recent works have tried to argue the expressivity of transformers with in-context learning. \citet{garg2022can,akyurek2023learning,fu2023transformers,ahn2024transformers,bai2024transformers,li2023transformersgenincontext,giannou2023looped,von2023uncovering,von2023transformers,panigrahi2023trainable,dai2023can} have argued that transformers can simulate specific machine learning algorithms (e.g. linear regression) on in-context examples.
However, the relation between the constructions and the performance of Transformers on real world datasets has been largely unclear.
Our framework shows that these constructions can be non-trivially extended to demonstrate how a Transformer could simulate a chatbot program.
A number of works have demonstrated the theoretical advantage of scratchpads~\citep{nye2021show} and chain-of-thought~\citep{wei2022chain} for the expressivity of bounded Transformer models~\citep{feng2024towards,li2024chain,nowak2024representational,merrill2024the,abbe2024far,hu2024limitation,hou2024universal}.
Our experiments illustrate how Transformers trained on ELIZA data make use of their own outputs to simulate data structures for dialogue tracking, highlighting the importance of intermediate outputs even without an explicit scratchpad.

\paragraph{Challenges for mechanistic interpretability} 
One direction for future work is to consider our ELIZA construction as a test bed for automatic interpretability methods---for example, compiling the construction into Transformer weights using Tracr~\citep{lindner2023tracr}.
Specifically, given a compiled Transformer corresponding to an ELIZA chatbot, to what extent could we recover the program using existing interpretability techniques, such as circuit finding~\citep{conmy2023towards,syed2023attribution} and dictionary learning~\citep{cunningham2023sparse,gurnee2024universal,marks2024sparse}?
Possible difficulties include sharing of attention heads across different ELIZA operations like parsing and copying, and sharing of mechanisms for different ELIZA operations like cycling and memory queues. 
As such, our framework might encourage more sophisticated interpretability techniques in the future.
Similarly, the ELIZA dataset could serve as a test-bed for recent approaches to designing intrinsically interpretable neural architectures for language tasks~\citep[e.g.][]{hewitt2023backpack,friedman2023learning}.

\paragraph{Mechanistic dependence on data}
Recent works have tried to understand the behavior of attention models trained on synthetic data. \citet{nanda2023progress} study feature formation in 1-layer transformer models on adders dataset. \citet{zhong2024clock} study the dependence on model hyperparameters and initialization. 
\citet{akyurek2024context,quirke2023training} study formation of $n$-gram induction heads in language models. \citet{allen2023physics,zhao2023transformers} study the behavior of LMs trained on different context-free grammars.  \citet{allen2023physics3point2,allen2024physics} further study knowledge manipulation and storage in LMs trained on synthetic datasets. \citet{zhang2022unveiling} propose LEGO synthetic reasoning dataset to understand generalization of transformers with simple boolean circuits. Finally, \citet{zhang2023trained,edelman2024evolution,nichani2024transformers} give end-to-end convergence analysis of self-attention models when trained under simplistic data assumptions.
ELIZA complements these studies by offering a rich but controlled setting requiring the composition of diverse subtasks.
Each subtask admits multiple plausible mechanisms, and, as shown in Section~\ref{sec:mechanisms}, different data distributional properties can lead to different mechanisms.
With increasing interest in formalizing the mechanistic relationship between data and training behavior \citep[e.g.][]{chan2022data,hahn2023theory,reddy2024mechanistic,xie2021explanation,jain2024mechanistically,lee2024mechanistic,prakash2024finetuning}, we believe ELIZA can be a useful test bed for future studies.

\section{Conclusion}
\label{sec:conclusion}
In this work, we constructed a Transformer that implements the classic ELIZA chatbot algorithm.
We then trained Transformers on ELIZA conversation transcripts and examined how well the models matched our construction.
Our constructions and dataset raise a number of possibilities for future research, including as a benchmark for automated interpretability methods, and as a setting for mechanistic analysis of learning dynamics.

\newpage

\paragraph{Acknowledgments}
We thank Adam Pearce, Adithya Bhaskar, Alexander Wettig, Howard Yen, and the members of the Princeton NLP group for helpful comments and discussion.
This research is funded by the National Science Foundation (IIS-2211779) and a Sloan Research Fellowship.
DF is supported by a Google PhD Fellowship.

\vspace{-0.1cm}
\paragraph{Limitations}
Our constructions illustrate some of the ways that Transformers could implement ELIZA, but they are not exhaustive, and they might not correspond to the solutions that Transformers actually learn.
Characterizing the mechanisms that models learn empirically is a key challenge for future work on interpretability.
Second, we conduct some analysis of the mechanisms that models learn, but we do not conduct an exhaustive mechanistic analysis; future work could conduct further analysis using other interpretability techniques, such as causal methods~\citep[e.g.][]{vig2020causal,feder2021causalm,geiger2021causal}, to understand how the mechanisms are encoded in the model's weights.
Third, we do not investigate whether open-source conversational models use similar mechanisms to the ones we considered here.
It is not straightforward to evaluate open-source conversational models on our synthetic task, because it is difficult to teach a model to follow the ELIZA algorithm and rules without further fine-tuning.
One possible direction for future work is to prompt an instruction-tuned model to follow the ELIZA rules and investigate which mechanisms it uses.

More generally, while ELIZA offers a setting for investigating a number of aspects of conversations, there is a substantial gap between ELIZA and real-world chatbots.
For example, ELIZA is a deterministic program, whereas most real-world chatbots are trained on data with more stochasticity.
One possible path for future research is to gradually extend the rule-based chatbot framework to include more of the key phenomena observed in modern language models, with the goal of understanding how these behaviors can be implemented with interpretable Transformer mechanisms.
These include more sophisticated pattern matching (for example, extending from regular expressions to semantic parsing); in-context learning; and explicit chain-of-thought reasoning.

\bibliography{ref}

\begin{thebibliography}{85}
\providecommand{\natexlab}[1]{#1}

\bibitem[{Abbe et~al.(2024)Abbe, Bengio, Lotfi, Sandon, and Saremi}]{abbe2024far}
Emmanuel Abbe, Samy Bengio, Aryo Lotfi, Colin Sandon, and Omid Saremi. 2024.
\newblock How far can transformers reason? {T}he locality barrier and inductive scratchpad.
\newblock \emph{arXiv preprint arXiv:2406.06467}.

\bibitem[{Ahn et~al.(2023)Ahn, Cheng, Daneshmand, and Sra}]{ahn2024transformers}
Kwangjun Ahn, Xiang Cheng, Hadi Daneshmand, and Suvrit Sra. 2023.
\newblock Transformers learn to implement preconditioned gradient descent for in-context learning.
\newblock \emph{Advances in Neural Information Processing Systems (NeurIPS)}, 36.

\bibitem[{Aky{\"u}rek et~al.(2023)Aky{\"u}rek, Schuurmans, Andreas, Ma, and Zhou}]{akyurek2023learning}
Ekin Aky{\"u}rek, Dale Schuurmans, Jacob Andreas, Tengyu Ma, and Denny Zhou. 2023.
\newblock What learning algorithm is in-context learning? {I}nvestigations with linear models.
\newblock In \emph{International Conference on Learning Representations (ICLR)}.

\bibitem[{Aky{\"u}rek et~al.(2024)Aky{\"u}rek, Wang, Kim, and Andreas}]{akyurek2024context}
Ekin Aky{\"u}rek, Bailin Wang, Yoon Kim, and Jacob Andreas. 2024.
\newblock In-context language learning: {A}rchitectures and algorithms.
\newblock \emph{arXiv preprint arXiv:2401.12973}.

\bibitem[{Allen-Zhu and Li(2023{\natexlab{a}})}]{allen2023physics}
Zeyuan Allen-Zhu and Yuanzhi Li. 2023{\natexlab{a}}.
\newblock Physics of language models: {P}art 1, {C}ontext-free grammar.
\newblock \emph{arXiv preprint arXiv:2305.13673}.

\bibitem[{Allen-Zhu and Li(2023{\natexlab{b}})}]{allen2023physics3point2}
Zeyuan Allen-Zhu and Yuanzhi Li. 2023{\natexlab{b}}.
\newblock Physics of language models: Part 3.2, {K}nowledge manipulation.
\newblock \emph{arXiv preprint arXiv:2309.14402}.

\bibitem[{Allen-Zhu and Li(2024)}]{allen2024physics}
Zeyuan Allen-Zhu and Yuanzhi Li. 2024.
\newblock Physics of language models: Part 3.3, {K}nowledge capacity scaling laws.
\newblock \emph{arXiv preprint arXiv:2404.05405}.

\bibitem[{Anil et~al.(2022)Anil, Wu, Andreassen, Lewkowycz, Misra, Ramasesh, Slone, Gur-Ari, Dyer, and Neyshabur}]{anil2022exploring}
Cem Anil, Yuhuai Wu, Anders Andreassen, Aitor Lewkowycz, Vedant Misra, Vinay Ramasesh, Ambrose Slone, Guy Gur-Ari, Ethan Dyer, and Behnam Neyshabur. 2022.
\newblock Exploring length generalization in large language models.
\newblock \emph{Advances in Neural Information Processing Systems (NeurIPS)}, 35:38546--38556.

\bibitem[{Bai et~al.(2024)Bai, Chen, Wang, Xiong, and Mei}]{bai2024transformers}
Yu~Bai, Fan Chen, Huan Wang, Caiming Xiong, and Song Mei. 2024.
\newblock Transformers as statisticians: {P}rovable in-context learning with in-context algorithm selection.
\newblock \emph{Advances in Neural Information Processing Systems (NeurIPS)}, 36.

\bibitem[{Beltagy et~al.(2020)Beltagy, Peters, and Cohan}]{beltagy2020longformer}
Iz~Beltagy, Matthew~E. Peters, and Arman Cohan. 2020.
\newblock Longformer: {T}he long-document transformer.
\newblock \emph{arXiv:2004.05150}.

\bibitem[{Bhattamishra et~al.(2020{\natexlab{a}})Bhattamishra, Ahuja, and Goyal}]{bhattamishra2020ability}
Satwik Bhattamishra, Kabir Ahuja, and Navin Goyal. 2020{\natexlab{a}}.
\newblock On the ability and limitations of transformers to recognize formal languages.
\newblock In \emph{Empirical Methods in Natural Language Processing (EMNLP)}, pages 7096--7116.

\bibitem[{Bhattamishra et~al.(2020{\natexlab{b}})Bhattamishra, Patel, and Goyal}]{bhattamishra2020computational}
Satwik Bhattamishra, Arkil Patel, and Navin Goyal. 2020{\natexlab{b}}.
\newblock On the computational power of {T}ransformers and its implications in sequence modeling.
\newblock In \emph{Computational Natural Language Learning (CoNLL)}, pages 455--475.

\bibitem[{Chan et~al.(2022)Chan, Santoro, Lampinen, Wang, Singh, Richemond, McClelland, and Hill}]{chan2022data}
Stephanie Chan, Adam Santoro, Andrew Lampinen, Jane Wang, Aaditya Singh, Pierre Richemond, James McClelland, and Felix Hill. 2022.
\newblock Data distributional properties drive emergent in-context learning in transformers.
\newblock \emph{Advances in Neural Information Processing Systems (NeurIPS)}, 35:18878--18891.

\bibitem[{Conmy et~al.(2023)Conmy, Mavor-Parker, Lynch, Heimersheim, and Garriga-Alonso}]{conmy2023towards}
Arthur Conmy, Augustine~N Mavor-Parker, Aengus Lynch, Stefan Heimersheim, and Adri{\`a} Garriga-Alonso. 2023.
\newblock Towards automated circuit discovery for mechanistic interpretability.
\newblock \emph{arXiv preprint arXiv:2304.14997}.

\bibitem[{Cunningham et~al.(2023)Cunningham, Ewart, Riggs, Huben, and Sharkey}]{cunningham2023sparse}
Hoagy Cunningham, Aidan Ewart, Logan Riggs, Robert Huben, and Lee Sharkey. 2023.
\newblock Sparse autoencoders find highly interpretable features in language models.
\newblock \emph{arXiv preprint arXiv:2309.08600}.

\bibitem[{Dai et~al.(2023)Dai, Sun, Dong, Hao, Ma, Sui, and Wei}]{dai2023can}
Damai Dai, Yutao Sun, Li~Dong, Yaru Hao, Shuming Ma, Zhifang Sui, and Furu Wei. 2023.
\newblock Why can {GPT} learn in-context? {L}anguage models secretly perform gradient descent as meta-optimizers.
\newblock In \emph{Findings of Association for Computational Linguistics (ACL)}.

\bibitem[{Edelman et~al.(2024)Edelman, Edelman, Goel, Malach, and Tsilivis}]{edelman2024evolution}
Benjamin~L Edelman, Ezra Edelman, Surbhi Goel, Eran Malach, and Nikolaos Tsilivis. 2024.
\newblock The evolution of statistical induction heads: {I}n-context learning {M}arkov chains.
\newblock \emph{arXiv preprint arXiv:2402.11004}.

\bibitem[{Edelman et~al.(2022)Edelman, Goel, Kakade, and Zhang}]{edelman2022inductive}
Benjamin~L Edelman, Surbhi Goel, Sham Kakade, and Cyril Zhang. 2022.
\newblock Inductive biases and variable creation in self-attention mechanisms.
\newblock In \emph{International Conference on Machine Learning (ICML)}, pages 5793--5831. PMLR.

\bibitem[{Feder et~al.(2021)Feder, Oved, Shalit, and Reichart}]{feder2021causalm}
Amir Feder, Nadav Oved, Uri Shalit, and Roi Reichart. 2021.
\newblock Causa{LM}: {C}ausal model explanation through counterfactual language models.
\newblock \emph{Computational Linguistics}, 47(2):333--386.

\bibitem[{Feng et~al.(2024)Feng, Zhang, Gu, Ye, He, and Wang}]{feng2024towards}
Guhao Feng, Bohang Zhang, Yuntian Gu, Haotian Ye, Di~He, and Liwei Wang. 2024.
\newblock Towards revealing the mystery behind chain of thought: {A} theoretical perspective.
\newblock \emph{Advances in Neural Information Processing Systems (NeurIPS)}, 36.

\bibitem[{Friedman et~al.(2023)Friedman, Wettig, and Chen}]{friedman2023learning}
Dan Friedman, Alexander Wettig, and Danqi Chen. 2023.
\newblock Learning {T}ransformer {P}rograms.
\newblock \emph{Advances in Neural Information Processing Systems}, 36.

\bibitem[{Fu et~al.(2023)Fu, Chen, Jia, and Sharan}]{fu2023transformers}
Deqing Fu, Tian-Qi Chen, Robin Jia, and Vatsal Sharan. 2023.
\newblock Transformers learn higher-order optimization methods for in-context learning: {A} study with linear models.
\newblock \emph{arXiv preprint arXiv:2310.17086}.

\bibitem[{Garg et~al.(2022)Garg, Tsipras, Liang, and Valiant}]{garg2022can}
Shivam Garg, Dimitris Tsipras, Percy~S Liang, and Gregory Valiant. 2022.
\newblock What can transformers learn in-context? {A} case study of simple function classes.
\newblock \emph{Advances in Neural Information Processing Systems (NeurIPS)}, 35:30583--30598.

\bibitem[{Geiger et~al.(2021)Geiger, Lu, Icard, and Potts}]{geiger2021causal}
Atticus Geiger, Hanson Lu, Thomas Icard, and Christopher Potts. 2021.
\newblock Causal abstractions of neural networks.
\newblock \emph{Advances in Neural Information Processing Systems (NeurIPS)}, 34:9574--9586.

\bibitem[{Giannou et~al.(2023)Giannou, Rajput, yong Sohn, Lee, Lee, and Papailiopoulos}]{giannou2023looped}
Angeliki Giannou, Shashank Rajput, Jy~yong Sohn, Kangwook Lee, Jason~D. Lee, and Dimitris Papailiopoulos. 2023.
\newblock Looped {T}ransformers as programmable computers.
\newblock In \emph{International Conference on Machine Learning (ICML)}.

\bibitem[{Gurnee et~al.(2024)Gurnee, Horsley, Guo, Kheirkhah, Sun, Hathaway, Nanda, and Bertsimas}]{gurnee2024universal}
Wes Gurnee, Theo Horsley, Zifan~Carl Guo, Tara~Rezaei Kheirkhah, Qinyi Sun, Will Hathaway, Neel Nanda, and Dimitris Bertsimas. 2024.
\newblock Universal neurons in {GPT}2 language models.
\newblock \emph{arXiv preprint arXiv:2401.12181}.

\bibitem[{Hahn(2020)}]{hahn2020theoretical}
Michael Hahn. 2020.
\newblock Theoretical limitations of self-attention in neural sequence models.
\newblock \emph{Transactions of the Association of Computational Linguistics (TACL)}, 8:156--171.

\bibitem[{Hahn and Goyal(2023)}]{hahn2023theory}
Michael Hahn and Navin Goyal. 2023.
\newblock A theory of emergent in-context learning as implicit structure induction.
\newblock \emph{arXiv preprint arXiv:2303.07971}.

\bibitem[{Hao et~al.(2022)Hao, Angluin, and Frank}]{hao2022formal}
Yiding Hao, Dana Angluin, and Robert Frank. 2022.
\newblock Formal language recognition by hard attention transformers: {P}erspectives from circuit complexity.
\newblock \emph{Transactions of the Association of Computational Linguistics (TACL)}, 10:800--810.

\bibitem[{Harris et~al.(2020)Harris, Millman, Van Der~Walt, Gommers, Virtanen, Cournapeau, Wieser, Taylor, Berg, Smith et~al.}]{harris2020array}
Charles~R Harris, K~Jarrod Millman, St{\'e}fan~J Van Der~Walt, Ralf Gommers, Pauli Virtanen, David Cournapeau, Eric Wieser, Julian Taylor, Sebastian Berg, Nathaniel~J Smith, et~al. 2020.
\newblock Array programming with {N}um{P}y.
\newblock \emph{Nature}, 585(7825):357--362.

\bibitem[{Haviv et~al.(2022)Haviv, Ram, Press, Izsak, and Levy}]{haviv2022transformer}
Adi Haviv, Ori Ram, Ofir Press, Peter Izsak, and Omer Levy. 2022.
\newblock Transformer language models without positional encodings still learn positional information.
\newblock In \emph{Findings of Empirical Methods in Natural Language Processing (EMNLP)}, pages 1382--1390.

\bibitem[{Hay and Millican(2022)}]{hay2022eliza}
Anthony Hay and Peter Millican. 2022.
\newblock {ELIZA} is {T}uring complete.
\newblock \url{https://sites.google.com/view/elizagen-org/blog/eliza-is-turing-complete}.
\newblock Accessed: 2024-01-09.

\bibitem[{Hewitt et~al.(2023)Hewitt, Thickstun, Manning, and Liang}]{hewitt2023backpack}
John Hewitt, John Thickstun, Christopher~D Manning, and Percy Liang. 2023.
\newblock Backpack language models.
\newblock In \emph{Association for Computational Linguistics (ACL)}, pages 9103--9125.

\bibitem[{Hou et~al.(2024)Hou, Brandfonbrener, Kakade, Jelassi, and Malach}]{hou2024universal}
Kaiying Hou, David Brandfonbrener, Sham Kakade, Samy Jelassi, and Eran Malach. 2024.
\newblock Universal length generalization with {T}uring {P}rograms.
\newblock \emph{arXiv preprint arXiv:2407.03310}.

\bibitem[{Hu et~al.(2024)Hu, Liu, and Jin}]{hu2024limitation}
Jiachen Hu, Qinghua Liu, and Chi Jin. 2024.
\newblock On limitation of transformer for learning {HMM}s.
\newblock \emph{arXiv preprint arXiv:2406.04089}.

\bibitem[{Jain et~al.(2024)Jain, Kirk, Lubana, Dick, Tanaka, Rockt{\"a}schel, Grefenstette, and Krueger}]{jain2024mechanistically}
Samyak Jain, Robert Kirk, Ekdeep~Singh Lubana, Robert~P. Dick, Hidenori Tanaka, Tim Rockt{\"a}schel, Edward Grefenstette, and David Krueger. 2024.
\newblock Mechanistically analyzing the effects of fine-tuning on procedurally defined tasks.
\newblock In \emph{International Conference on Learning Representations (ICLR)}.

\bibitem[{Jurafsky and Martin(2020)}]{jurafsky2020chatbots}
Daniel Jurafsky and James~H Martin. 2020.
\newblock Chatbots and dialogue systems.
\newblock \emph{Speech and Language Processing}.

\bibitem[{Kazemnejad et~al.(2023)Kazemnejad, Padhi, Natesan, Das, and Reddy}]{kazemnejad2023impact}
Amirhossein Kazemnejad, Inkit Padhi, Karthikeyan Natesan, Payel Das, and Siva Reddy. 2023.
\newblock The impact of positional encoding on length generalization in {T}ransformers.
\newblock In \emph{Advances in Neural Information Processing Systems (NeurIPS)}.

\bibitem[{Kingma and Ba(2014)}]{kingma2014adam}
Diederik~P Kingma and Jimmy Ba. 2014.
\newblock Adam: {A} method for stochastic optimization.
\newblock \emph{arXiv preprint arXiv:1412.6980}.

\bibitem[{Lee et~al.(2024)Lee, Bai, Pres, Wattenberg, Kummerfeld, and Mihalcea}]{lee2024mechanistic}
Andrew Lee, Xiaoyan Bai, Itamar Pres, Martin Wattenberg, Jonathan~K Kummerfeld, and Rada Mihalcea. 2024.
\newblock A mechanistic understanding of alignment algorithms: {A} case study on {DPO} and toxicity.
\newblock \emph{International Conference on Machine Learning (ICML)}.

\bibitem[{Li et~al.(2023)Li, Ildiz, Papailiopoulos, and Oymak}]{li2023transformersgenincontext}
Yingcong Li, Muhammed~Emrullah Ildiz, Dimitris Papailiopoulos, and Samet Oymak. 2023.
\newblock Transformers as algorithms: {G}eneralization and stability in in-context learning.
\newblock In \emph{International Conference on Machine Learning (ICML)}, pages 19565--19594. PMLR.

\bibitem[{Li et~al.(2024)Li, Liu, Zhou, and Ma}]{li2024chain}
Zhiyuan Li, Hong Liu, Denny Zhou, and Tengyu Ma. 2024.
\newblock Chain of thought empowers transformers to solve inherently serial problems.
\newblock \emph{arXiv preprint arXiv:2402.12875}.

\bibitem[{Lindner et~al.(2023)Lindner, Kram{\'a}r, Farquhar, Rahtz, McGrath, and Mikulik}]{lindner2023tracr}
David Lindner, J{\'a}nos Kram{\'a}r, Sebastian Farquhar, Matthew Rahtz, Tom McGrath, and Vladimir Mikulik. 2023.
\newblock Tracr: {C}ompiled transformers as a laboratory for interpretability.
\newblock \emph{Advances in Neural Information Processing Systems (NeurIPS)}, 36.

\bibitem[{Liu et~al.(2023)Liu, Ash, Goel, Krishnamurthy, and Zhang}]{liu2023transformers}
Bingbin Liu, Jordan~T. Ash, Surbhi Goel, Akshay Krishnamurthy, and Cyril Zhang. 2023.
\newblock Transformers learn shortcuts to automata.
\newblock In \emph{International Conference on Learning Representations (ICLR)}.

\bibitem[{Marks et~al.(2024)Marks, Rager, Michaud, Belinkov, Bau, and Mueller}]{marks2024sparse}
Samuel Marks, Can Rager, Eric~J Michaud, Yonatan Belinkov, David Bau, and Aaron Mueller. 2024.
\newblock Sparse feature circuits: {D}iscovering and editing interpretable causal graphs in language models.
\newblock \emph{arXiv preprint arXiv:2403.19647}.

\bibitem[{McNaughton and Papert(1971)}]{mcnaughton1971counter}
Robert McNaughton and Seymour~A Papert. 1971.
\newblock \emph{Counter-Free Automata ({MIT} research monograph no. 65)}.
\newblock The MIT Press.

\bibitem[{Merrill and Sabharwal(2023)}]{merrill2023logic}
William Merrill and Ashish Sabharwal. 2023.
\newblock A logic for expressing log-precision transformers.
\newblock In \emph{Advances in Neural Information Processing Systems (NeurIPS)}.

\bibitem[{Merrill and Sabharwal(2024)}]{merrill2024the}
William Merrill and Ashish Sabharwal. 2024.
\newblock The expressive power of {T}ransformers with chain of thought.
\newblock In \emph{International Conference on Learning Representations (ICLR)}.

\bibitem[{Merrill et~al.(2022)Merrill, Sabharwal, and Smith}]{merrill2022saturated}
William Merrill, Ashish Sabharwal, and Noah~A Smith. 2022.
\newblock Saturated {T}ransformers are constant-depth threshold circuits.
\newblock \emph{Transactions of the Association of Computational Linguistics (TACL)}, 10:843--856.

\bibitem[{Nanda et~al.(2023)Nanda, Chan, Lieberum, Smith, and Steinhardt}]{nanda2023progress}
Neel Nanda, Lawrence Chan, Tom Lieberum, Jess Smith, and Jacob Steinhardt. 2023.
\newblock Progress measures for grokking via mechanistic interpretability.
\newblock In \emph{International Conference on Learning Representations (ICLR)}.

\bibitem[{Nichani et~al.(2024)Nichani, Damian, and Lee}]{nichani2024transformers}
Eshaan Nichani, Alex Damian, and Jason~D Lee. 2024.
\newblock How transformers learn causal structure with gradient descent.
\newblock \emph{arXiv preprint arXiv:2402.14735}.

\bibitem[{Nowak et~al.(2024)Nowak, Svete, Butoi, and Cotterell}]{nowak2024representational}
Franz Nowak, Anej Svete, Alexandra Butoi, and Ryan Cotterell. 2024.
\newblock On the representational capacity of neural language models with chain-of-thought reasoning.
\newblock \emph{arXiv preprint arXiv:2406.14197}.

\bibitem[{Nye et~al.(2021)Nye, Andreassen, Gur-Ari, Michalewski, Austin, Bieber, Dohan, Lewkowycz, Bosma, Luan et~al.}]{nye2021show}
Maxwell Nye, Anders~Johan Andreassen, Guy Gur-Ari, Henryk Michalewski, Jacob Austin, David Bieber, David Dohan, Aitor Lewkowycz, Maarten Bosma, David Luan, et~al. 2021.
\newblock Show your work: {S}cratchpads for intermediate computation with language models.
\newblock \emph{arXiv preprint arXiv:2112.00114}.

\bibitem[{Olsson et~al.(2022)Olsson, Elhage, Nanda, Joseph, DasSarma, Henighan, Mann, Askell, Bai, Chen et~al.}]{olsson2022context}
Catherine Olsson, Nelson Elhage, Neel Nanda, Nicholas Joseph, Nova DasSarma, Tom Henighan, Ben Mann, Amanda Askell, Yuntao Bai, Anna Chen, et~al. 2022.
\newblock In-context learning and induction heads.
\newblock \emph{arXiv preprint arXiv:2209.11895}.

\bibitem[{Panigrahi et~al.(2023)Panigrahi, Malladi, Xia, and Arora}]{panigrahi2023trainable}
Abhishek Panigrahi, Sadhika Malladi, Mengzhou Xia, and Sanjeev Arora. 2023.
\newblock Trainable transformer in transformer.
\newblock \emph{arXiv preprint arXiv:2307.01189}.

\bibitem[{Paszke et~al.(2019)Paszke, Gross, Massa, Lerer, Bradbury, Chanan, Killeen, Lin, Gimelshein, Antiga et~al.}]{paszke2019pytorch}
Adam Paszke, Sam Gross, Francisco Massa, Adam Lerer, James Bradbury, Gregory Chanan, Trevor Killeen, Zeming Lin, Natalia Gimelshein, Luca Antiga, et~al. 2019.
\newblock Py{T}orch: {A}n imperative style, high-performance deep learning library.
\newblock \emph{Advances in Neural Information Processing Systems (NeurIPS)}, 32.

\bibitem[{P{\'e}rez et~al.(2021)P{\'e}rez, Barcel{\'o}, and Marinkovic}]{perez2021attention}
Jorge P{\'e}rez, Pablo Barcel{\'o}, and Javier Marinkovic. 2021.
\newblock Attention is {T}uring-complete.
\newblock \emph{The Journal of Machine Learning Research (JMLR)}, 22(75):1--35.

\bibitem[{Pin(2020)}]{pin2020prove}
Jean-{\'E}ric Pin. 2020.
\newblock How to prove that a language is regular or star-free?
\newblock In \emph{International Conference on Language and Automata Theory and Applications}, pages 68--88.

\bibitem[{Prakash et~al.(2024)Prakash, Shaham, Haklay, Belinkov, and Bau}]{prakash2024finetuning}
Nikhil Prakash, Tamar~Rott Shaham, Tal Haklay, Yonatan Belinkov, and David Bau. 2024.
\newblock Fine-tuning enhances existing mechanisms: {A} case study on entity tracking.
\newblock In \emph{International Conference on Learning Representations (ICLR)}.

\bibitem[{Pérez et~al.(2019)Pérez, Marinković, and Barceló}]{perez2019turing}
Jorge Pérez, Javier Marinković, and Pablo Barceló. 2019.
\newblock On the {T}uring completeness of modern neural network architectures.
\newblock In \emph{International Conference on Learning Representations (ICLR)}.

\bibitem[{Quirke et~al.(2023)Quirke, Heindrich, Gurnee, and Nanda}]{quirke2023training}
Lucia Quirke, Lovis Heindrich, Wes Gurnee, and Neel Nanda. 2023.
\newblock Training dynamics of contextual n-grams in language models.
\newblock \emph{arXiv preprint arXiv:2311.00863}.

\bibitem[{Radford et~al.(2019)Radford, Wu, Child, Luan, Amodei, Sutskever et~al.}]{radford2019language}
Alec Radford, Jeffrey Wu, Rewon Child, David Luan, Dario Amodei, Ilya Sutskever, et~al. 2019.
\newblock Language models are unsupervised multitask learners.
\newblock \emph{OpenAI blog}, 1(8):9.

\bibitem[{Reddy(2024)}]{reddy2024mechanistic}
Gautam Reddy. 2024.
\newblock The mechanistic basis of data dependence and abrupt learning in an in-context classification task.
\newblock In \emph{International Conference on Learning Representations (ICLR)}.

\bibitem[{Singh et~al.(2024)Singh, Moskovitz, Hill, Chan, and Saxe}]{singh2024needs}
Aaditya~K Singh, Ted Moskovitz, Felix Hill, Stephanie~CY Chan, and Andrew~M Saxe. 2024.
\newblock What needs to go right for an induction head? {A} mechanistic study of in-context learning circuits and their formation.
\newblock \emph{arXiv preprint arXiv:2404.07129}.

\bibitem[{Strobl et~al.(2024)Strobl, Merrill, Weiss, Chiang, and Angluin}]{strobl2024formal}
Lena Strobl, William Merrill, Gail Weiss, David Chiang, and Dana Angluin. 2024.
\newblock What formal languages can {T}ransformers express? {A} survey.
\newblock \emph{Transactions of the Association of Computational Linguistics (TACL)}, 12:543--561.

\bibitem[{Svete and Cotterell(2024)}]{svete2024transformers}
Anej Svete and Ryan Cotterell. 2024.
\newblock Transformers can represent $n$-gram language models.
\newblock In \emph{North American Chapter of the Association for Computational Linguistics: Human Language Technologies (NAACL-HLT)}, pages 6841--6874.

\bibitem[{Syed et~al.(2023)Syed, Rager, and Conmy}]{syed2023attribution}
Aaquib Syed, Can Rager, and Arthur Conmy. 2023.
\newblock Attribution patching outperforms automated circuit discovery.
\newblock \emph{arXiv preprint arXiv:2310.10348}.

\bibitem[{Vaswani et~al.(2017)Vaswani, Shazeer, Parmar, Uszkoreit, Jones, Gomez, Kaiser, and Polosukhin}]{vaswani2017attention}
Ashish Vaswani, Noam Shazeer, Niki Parmar, Jakob Uszkoreit, Llion Jones, Aidan~N Gomez, {\L}ukasz Kaiser, and Illia Polosukhin. 2017.
\newblock Attention is all you need.
\newblock \emph{Advances in Neural Information Processing Systems (NeurIPS)}, 30.

\bibitem[{Vig et~al.(2020)Vig, Gehrmann, Belinkov, Qian, Nevo, Sakenis, Huang, Singer, and Shieber}]{vig2020causal}
Jesse Vig, Sebastian Gehrmann, Yonatan Belinkov, Sharon Qian, Daniel Nevo, Simas Sakenis, Jason Huang, Yaron Singer, and Stuart Shieber. 2020.
\newblock Causal mediation analysis for interpreting neural {NLP}: {T}he case of gender bias.
\newblock \emph{arXiv preprint arXiv:2004.12265}.

\bibitem[{Von~Oswald et~al.(2023)Von~Oswald, Niklasson, Randazzo, Sacramento, Mordvintsev, Zhmoginov, and Vladymyrov}]{von2023transformers}
Johannes Von~Oswald, Eyvind Niklasson, Ettore Randazzo, Jo{\~a}o Sacramento, Alexander Mordvintsev, Andrey Zhmoginov, and Max Vladymyrov. 2023.
\newblock Transformers learn in-context by gradient descent.
\newblock In \emph{International Conference on Machine Learning (ICML)}, pages 35151--35174. PMLR.

\bibitem[{von Oswald et~al.(2023)von Oswald, Niklasson, Schlegel, Kobayashi, Zucchet, Scherrer, Miller, Sandler, Vladymyrov, Pascanu et~al.}]{von2023uncovering}
Johannes von Oswald, Eyvind Niklasson, Maximilian Schlegel, Seijin Kobayashi, Nicolas Zucchet, Nino Scherrer, Nolan Miller, Mark Sandler, Max Vladymyrov, Razvan Pascanu, et~al. 2023.
\newblock Uncovering mesa-optimization algorithms in transformers.
\newblock \emph{arXiv preprint arXiv:2309.05858}.

\bibitem[{Waskom(2021)}]{waskom2021seaborn}
Michael~L Waskom. 2021.
\newblock Seaborn: {S}tatistical data visualization.
\newblock \emph{Journal of Open Source Software}, 6(60):3021.

\bibitem[{Wei et~al.(2022{\natexlab{a}})Wei, Chen, and Ma}]{wei2022statistically}
Colin Wei, Yining Chen, and Tengyu Ma. 2022{\natexlab{a}}.
\newblock Statistically meaningful approximation: {A} case study on approximating {T}uring {M}achines with {T}ransformers.
\newblock \emph{Advances in Neural Information Processing Systems (NeurIPS)}, 35:12071--12083.

\bibitem[{Wei et~al.(2022{\natexlab{b}})Wei, Wang, Schuurmans, Bosma, Xia, Chi, Le, Zhou et~al.}]{wei2022chain}
Jason Wei, Xuezhi Wang, Dale Schuurmans, Maarten Bosma, Fei Xia, Ed~Chi, Quoc~V Le, Denny Zhou, et~al. 2022{\natexlab{b}}.
\newblock Chain-of-{T}hought prompting elicits reasoning in large language models.
\newblock \emph{Advances in Neural Information Processing Systems (NeurIPS)}, 35:24824--24837.

\bibitem[{Weiss et~al.(2021)Weiss, Goldberg, and Yahav}]{weiss2021thinking}
Gail Weiss, Yoav Goldberg, and Eran Yahav. 2021.
\newblock Thinking like {T}ransformers.
\newblock In \emph{International Conference on Machine Learning (ICML)}, pages 11080--11090. PMLR.

\bibitem[{Weizenbaum(1966)}]{weizenbaum1966eliza}
Joseph Weizenbaum. 1966.
\newblock {ELIZA}---{A} computer program for the study of natural language communication between man and machine.
\newblock \emph{Communications of the ACM}, 9(1):36--45.

\bibitem[{Wolf et~al.(2020)Wolf, Debut, Sanh, Chaumond, Delangue, Moi, Cistac, Rault, Louf, Funtowicz et~al.}]{wolf2020transformers}
Thomas Wolf, Lysandre Debut, Victor Sanh, Julien Chaumond, Clement Delangue, Anthony Moi, Pierric Cistac, Tim Rault, R{\'e}mi Louf, Morgan Funtowicz, et~al. 2020.
\newblock Transformers: {S}tate-of-the-art natural language processing.
\newblock In \emph{Empirical Methods in Natural Language Processing (EMNLP): System Demonstrations}, pages 38--45.

\bibitem[{Xie et~al.(2021)Xie, Raghunathan, Liang, and Ma}]{xie2021explanation}
Sang~Michael Xie, Aditi Raghunathan, Percy Liang, and Tengyu Ma. 2021.
\newblock An explanation of in-context learning as implicit {B}ayesian inference.
\newblock \emph{arXiv preprint arXiv:2111.02080}.

\bibitem[{Yang et~al.(2024)Yang, Chiang, and Angluin}]{yang2024masked}
Andy Yang, David Chiang, and Dana Angluin. 2024.
\newblock Masked hard-attention transformers recognize exactly the star-free languages.
\newblock In \emph{Advances in Neural Information Processing Systems (NeurIPS)}.

\bibitem[{Yao et~al.(2021)Yao, Peng, Papadimitriou, and Narasimhan}]{yao2021self}
Shunyu Yao, Binghui Peng, Christos Papadimitriou, and Karthik Narasimhan. 2021.
\newblock Self-attention networks can process bounded hierarchical languages.
\newblock In \emph{Association for Computational Linguistics and International Joint Conference on Natural Language Processing (ACL-IJCNLP)}, pages 3770--3785.

\bibitem[{Zhang et~al.(2023)Zhang, Frei, and Bartlett}]{zhang2023trained}
Ruiqi Zhang, Spencer Frei, and Peter~L Bartlett. 2023.
\newblock Trained transformers learn linear models in-context.
\newblock \emph{arXiv preprint arXiv:2306.09927}.

\bibitem[{Zhang et~al.(2022)Zhang, Backurs, Bubeck, Eldan, Gunasekar, and Wagner}]{zhang2022unveiling}
Yi~Zhang, Arturs Backurs, S{\'e}bastien Bubeck, Ronen Eldan, Suriya Gunasekar, and Tal Wagner. 2022.
\newblock Unveiling {T}ransformers with {LEGO}: {A} synthetic reasoning task.
\newblock \emph{arXiv preprint arXiv:2206.04301}.

\bibitem[{Zhao et~al.(2023)Zhao, Panigrahi, Ge, and Arora}]{zhao2023transformers}
Haoyu Zhao, Abhishek Panigrahi, Rong Ge, and Sanjeev Arora. 2023.
\newblock Do transformers parse while predicting the masked word?
\newblock In \emph{Empirical Methods in Natural Language Processing (EMNLP)}, pages 16513--16542.

\bibitem[{Zhong et~al.(2023)Zhong, Liu, Tegmark, and Andreas}]{zhong2024clock}
Ziqian Zhong, Ziming Liu, Max Tegmark, and Jacob Andreas. 2023.
\newblock The clock and the pizza: {T}wo stories in mechanistic explanation of neural networks.
\newblock \emph{Advances in Neural Information Processing Systems (NeurIPS)}, 36.

\bibitem[{Zhou et~al.(2024)Zhou, Bradley, Littwin, Razin, Saremi, Susskind, Bengio, and Nakkiran}]{zhou2023algorithms}
Hattie Zhou, Arwen Bradley, Etai Littwin, Noam Razin, Omid Saremi, Josh Susskind, Samy Bengio, and Preetum Nakkiran. 2024.
\newblock What algorithms can {T}ransformers learn? {A} study in length generalization.
\newblock In \emph{International Conference on Learning Representations (ICLR)}.

\end{thebibliography}

\newpage
\appendix

\begin{table*}
    \centering
    \small
\begin{tabular}{>{\hangindent=1em}p{0.95\linewidth}}
Men are all alike.\\
IN WHAT WAY\\
They’re always bugging us about something or other.\\
CAN YOU THINK OF A SPECIFIC EXAMPLE\\
Well, my boyfriend made me come here.\\
YOUR BOYFRIEND MADE YOU COME HERE\\
He says I’m depressed much of the time.\\
I AM SORRY TO HEAR YOU ARE DEPRESSED\\
It’s true. I am unhappy.\\
DO YOU THINK COMING HERE WILL HELP YOU NOT TO BE UNHAPPY\\
I need some help, that much seems certain.\\
WHAT WOULD IT MEAN TO YOU IF YOU GOT SOME HELP\\
Perhaps I could learn to get along with my mother.\\
TELL ME MORE ABOUT YOUR FAMILY\\
My mother takes care of me.\\
WHO ELSE IN YOUR FAMILY TAKES CARE OF YOU\\
My father.\\
YOUR FATHER\\
You are like my father in some ways.\\
WHAT RESEMBLANCE DO YOU SEE\\
You are not very aggressive, but I think you don’t want me to notice that.\\
WHAT MAKES YOU THINK I AM NOT VERY AGGRESSIVE\\
You don’t argue with me.\\
WHY DO YOU THINK I DON’T ARGUE WITH YOU\\
You are afraid of me.\\
DOES IT PLEASE YOU TO BELIEVE I AM AFRAID OF YOU\\
My father is afraid of everybody.\\
WHAT ELSE COMES TO MIND WHEN YOU THINK OF YOUR FATHER\\
Bullies.\\
DOES THAT HAVE ANYTHING TO DO WITH THE FACT THAT YOUR BOYFRIEND MADE YOU COME HERE    \\
    \end{tabular}
\caption{
An example ELIZA conversation, reproduced from~\citet{weizenbaum1966eliza}.
Lines with all capital letters are generated by ELIZA.
\label{tab:eliza_example}
}
\end{table*}

\section{ELIZA Algorithm Details}
\label{app:eliza_details}

Here we provide some additional details about the ELIZA algorithm.
Our presentation of the ELIZA algorithm in Section~\ref{sec:background} omits some details of the original ELIZA algorithm, to improve clarity, so we describe these details here.\footnote{
For an annotated explanation of an ELIZA script, see \url{https://github.com/jeffshrager/elizagen.org/blob/master/1965_Weizenbaum_MAD-SLIP/1966_01_CACM_article_Eliza_script.txt}.
For more resources related to ELIZA, see \url{http://elizagen.org/}.
}

\paragraph{Word-level translation}
An ELIZA script can include word-level translation rules---for example, \texttt{I = YOU}, \texttt{YOU = I}, and \texttt{ME = YOU}.
These translations are applied to all of the words in the input before trying to match the input to a pattern.
Therefore, in the original ELIZA script, the patterns are written to match inputs after the word-level translations have been applied.
So, for example, the rule
\begin{align*}
    \texttt{0 ARE I 0} \to \text{Would you prefer if I weren't 4?}
\end{align*}
would match the input ``Are you laughing at me?'' and transform it to ``Would you prefer if I weren't laughing at you?''
In this paper, we write rules to match the input prior to word-level translations---so, for example, we would present the pattern above as \texttt{0 ARE YOU 0}.
Word-level translation is trivial to incorporate into the Transformer construction, by using the final linear layer to map each word to its translation.

\begin{figure*}[t]
\centering
\includegraphics[width=0.6\textwidth]{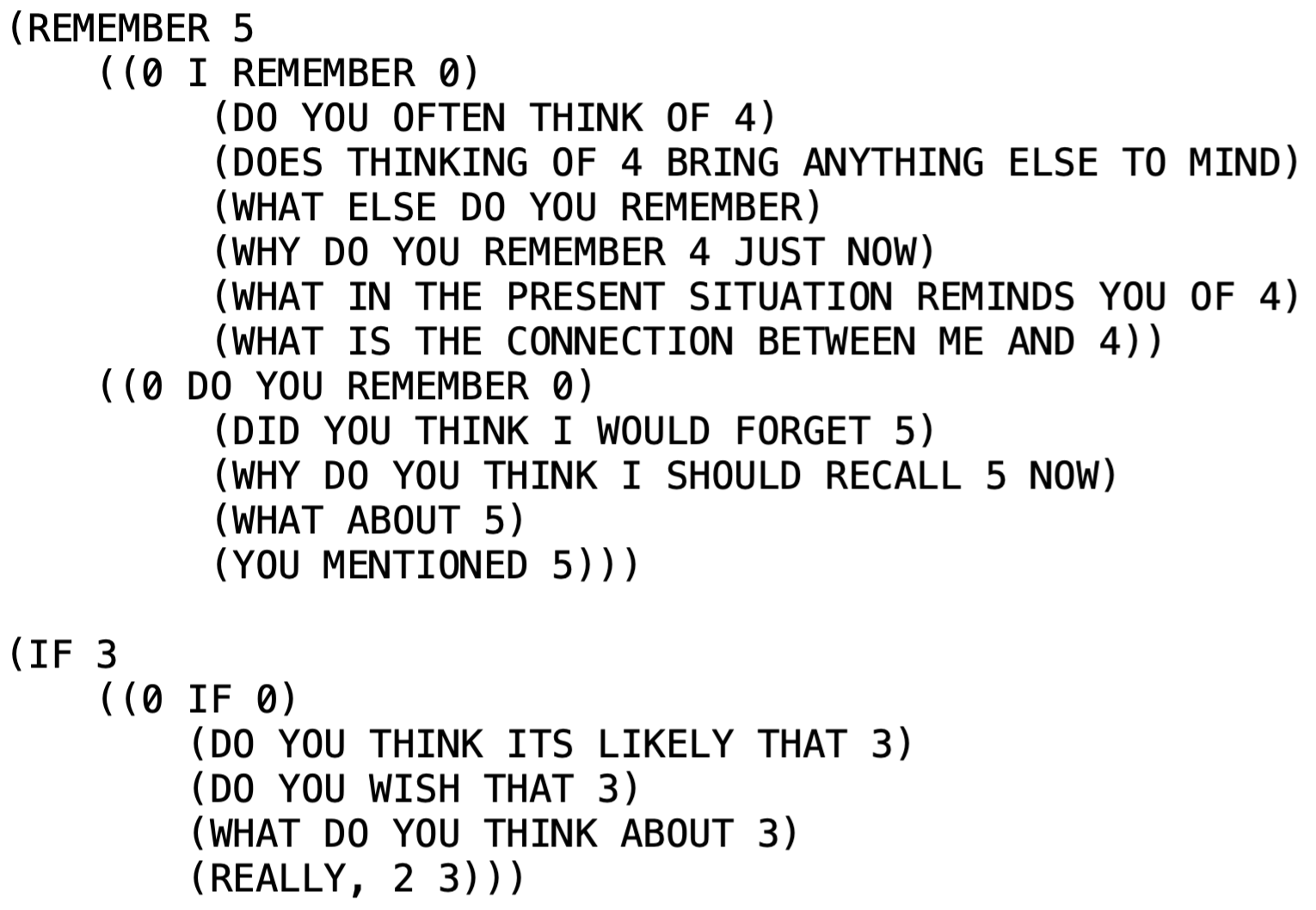}
\caption{Part of an ELIZA script, from~\citet{weizenbaum1966eliza}.
Each entry in the script consists of a ranked keyword and a list of patterns, with each pattern associated with multiple transformation rules.}
\label{fig:eliza_script_example}
\end{figure*}

\paragraph{Keywords}
Each entry in an ELIZA script consists of a ranked keyword.
Each keyword is associated with a list of decomposition templates, and each decomposition template is associated with one or more transformation rules.
See Figure~\ref{fig:eliza_script_example} for an example.
To select a decomposition template, ELIZA finds the highest ranked keyword that appears in the input, and then finds the first decomposition template in the associated list that matches the input.
If none of the templates matched, ELIZA checks the next highest-ranked keyword.
In this paper, we ignore the role of keywords, and instead define an ELIZA script by a set of ranked decomposition templates and associated transformation rules.

\paragraph{Pre-transformation rules}
The pre-transformation rule is a special rule that applies a transformation to the input, and then ``passes control'' to another keyword in the script.
There is one use of the pre-transformation rule in Weizenbaum's ELIZA script: if the input matches the pattern \ttt{0 I'm 0}, it is reassembled as ``I am 3,'' and then matched against templates with the keyword ``am,'' such as \ttt{0 I am 0}.
However, the pre-transformation rule is critical to the construction of~\citet{hay2022eliza} for embedding a Turing machine in an ELIZA script, which we will discuss in more detail below (App~\ref{app:turing_machine}).
In this construction, the input at each step represents the tape of the Turing machine, and keywords in the script correspond to states.
Each pre-transformation rule transforms the input by applying one update to the tape, and then passes control to a new keyword corresponding to the next state.

\section{Construction Details}
\label{app:construction_details}

In this section, we provide additional details about our ELIZA constructions, including sample implementations in RASP~\citep{weiss2021thinking}.
The input to a RASP program is a sequence of \ttt{tokens}.
The program then consists of a series of operations that output new sequences of equal length to \ttt{tokens}, corresponding to intermediate embeddings in the Transformer.
The \ttt{select} and \ttt{aggregate} operations correspond to the attention mechanism in the Transformer; these are the only operations that can combine information from different positions in the sequence.
All other operations must operate independently at each position, corresponding to feedforward layers.
Like~\citet{weiss2021thinking}, we allow feedforward layers to implement arbitrary element-wise transformations.
We do not provide explicit constructions for these element-wise transformations; we leave this for future work.
Figure~\ref{fig:rasp} shows the RASP~\citep{weiss2021thinking} attention primitives we use in our construction, implemented in NumPy~\citep{harris2020array}.

\begin{figure*}[t]
\begin{minted}[fontsize=\scriptsize]{python}
def select(keys, queries, predicate):
    # Calculate a (binary) attention pattern.
    selector = np.array([[predicate(q, k) for k in keys] for q in queries])
    return np.tril(selector)

def selector_width(selector, max_width=None):
    # Count the number of keys attended by each query, up to `max_width`.
    width = selector.sum(-1)
    if max_width:
        return np.minimum(width, max_width)
    return width

def aggregate(selector, values, one_hot=False):
    # Aggregate either a single value vector or a batch of value vectors
    # stored in a dictionary.
    if type(values) == dict:
        return {k: aggregate(selector, v, one_hot) for k, v in values.items()}
    if one_hot:
        return values[selector.argmax(-1)]
    attn = selector / np.maximum(selector.sum(-1, keepdims=True), 1e-9)
    return values @ attn.T
\end{minted}
\caption{
Code for the  primitive RASP operations~\citep{weiss2021thinking} we use in our construction, using NumPy~\citep{harris2020array}.
Each attention head can implement one pair of \ttt{select} and \ttt{aggregate} operations.
The \ttt{selector\_width} function corresponds to an attention head followed by a feed-forward layer, which maps the scalar attention output to an embedding that can be used in subsequent attention layers.
Because \ttt{selector\_width} maps each possible width to a unique, orthogonal embedding, the program must specify in advance the maximum width it will handle.
}
\label{fig:rasp}
\end{figure*}

\subsection{Input Segmentation and Position Encoding}
\label{app:input_segmentation}
Our first step is to divide the input into segments, corresponding to the turns in the conversation. 
This is accomplished by using the special delimiter tokens to count the number of utterances seen so far:
\begin{minted}[fontsize=\scriptsize]{python}
segment_ids = selector_width(
  select(tokens, tokens, lambda q, k: k in ("u:", "e:"),
  max_width=max_segments)
\end{minted}
We will use these \ttt{segment\_ids} throughout the construction to restrict attention to a particular utterance.
The \ttt{segment\_ids} are also used to generate local positional encodings:
\begin{minted}[fontsize=\scriptsize]{python}
segment_positions = selector_width(
  select(segment_ids, segment_ids, ==),
  max_width=max_segment_length)
\end{minted}
This value encodes the position relative to the start of the current segment.

\paragraph{Remark on length generalization}
While not the focus of our investigation here, our approach to segment and position encodings has implications for length generalization, similar to the cases studied by~\citet{zhou2023algorithms}.
In particular, we must specify in advance the maximum number of segments per conversation, as well as the length of each segment.
This is because the \ttt{selector\_width} operator is implemented using one attention layer followed by one feed-forward layer.
At each position $i$, the attention layer outputs $1/c$, where $c$ is the number of key positions attended to from position $i$.
The feed-forward layer then maps each value of $1/c$ to an orthogonal embedding.
In our construction, we implement this second step as a look-up table, meaning that we must decide in advance on the maximum possible value of $c$.
This means that our construction sets a limit on the number of segments per conversation, as well as the length within each segment.
If a model learned this mechanism, we would expect it to fail to generalize if the number of segments or the length of a segment increases beyond the training set.
(On the other hand, the construction does not place a direct limit on the total conversation length.)

\subsection{Template Matching}
\label{app:template_matching}

\begin{figure*}[t]
\begin{minted}[fontsize=\scriptsize]{python}
def match_templates(tokens, segment_ids, segment_positions, templates):
  L = max(len(t) for t in templates)
  prefixes = [{("u:",): tokens == "u:"}]
  
  # Each layer l checks if the input matches t[:l+1]
  for l in range(1, L):
    just_matched = select_prev(prefixes[-1], segment_ids, segment_positions)
    ever_matched = frac_prev(prefixes[-1], segment_ids, segment_positions)
    new_matches = {}
    for t in templates:
      if len(t) <= l: continue
      if t[l] == "0":
        new_matches[t[:l+1]] = ever_matched[t[:l]] > 0
      elif t[l-1] == "0" and t[l] == "1":
        new_matches[t[:l+1]] = prefixes[-1][t[:l]]
      elif t[l-1] == "0":
        new_matches[t[:l+1]] = prefixes[-1][t[:l]] & (tokens == t[l])
      elif t[l] == "1":
        new_matches[t[:l+1]] = just_matched[t[:l]]
      else:
        new_matches[t[:l+1]] = just_matched[t[:l]] & (tokens == t[l])
    prefixes.append(new_matches)
    
  # For each template, identify the longest matching prefix at each position.
  states = {}
  for t in templates:
    s = np.stack([m[t[:l+1]] for l, m in zip(range(len(t)), prefixes)])
    ind = np.arange(s.shape[0])
    states[t] = (ind[:, None] * s).max(0)
      
  return states
\end{minted}
\caption{
Code for matching an input sequence $\ttt{tokens}$ to a set of decomposition templates.
}
\label{fig:match_templates}
\end{figure*}

The next step in the construction is to compare the most recent input to the inventory of decomposition templates.
Template matching involves two things: finding a template that matches the input, and decomposing the input according to that template's decomposition groups.
Our construction makes use of the fact that ELIZA templates are equivalent to star-free regular expressions~\citep{mcnaughton1971counter,pin2020prove}.
As a result, we can recognize these by simulating the corresponding finite-state automaton, building on the constructions of~\citet{liu2023transformers} and~\citet{yang2024masked}, adapted to recognize multiple templates in parallel.

\paragraph{Decomposition templates}
Given a vocabulary $\gV$, a decomposition template is a sequence $t = t_1, \ldots, t_L$, where each $t_i$ is either a word from $\gV$; the wildcard character $\ttt{0}$, which matches a sequence of zero or more words from $\gV$; or a positive integer $n$, which matches a sequence of exactly $n$ words from $\gV$.\footnote{
A template can also include an equivalence class $W \subset \gV$, which matches one instance of any word in $W$.
For example, the template \ttt{1(a|b)1} matches both \ttt{cab} and \ttt{cbb}.
This can be addressed at the embedding layer by assigning one dimension to the value of the indicator $\mathbf{1}\{w \in \gW\}$ for each word $w$.
}
We assume that the vocabulary contains two special beginning- and end-of-sequence delimiters, $\ttt{\^{}}$ and $\ttt{\$}$ respectively, and for every input $w_1, \ldots, w_N$ and template $t_1, \ldots, t_L$, $w_1 = t_1 = \ttt{\^{}}$ and $w_N = t_L = \ttt{\$}$.
We will use $t_{:i}$ to denote the template prefix $t_1, \ldots, t_i$.
As a working example, consider the vocabulary $\gV = \{\ttt{a}, \ttt{b}\}$ and the template $t = \ttt{\^{}a0bb0\$}$.
This template matches the input \ttt{\^{}aaabbaa\$} and decomposes it into five groups:
$\text{(1)} \; \ttt{a}
\; \text{(2)} \; \ttt{aa}
\; \text{(3)} \; \ttt{b}
\; \text{(4)} \; \ttt{b}
\; \text{(5)} \; \ttt{aa}$.
We always take a greedy approach to template matching: for example, using the same template, the input \ttt{\^{}aabbbaa\$} will be decomposed as $\text{(1)} \; \ttt{a}
\; \text{(2)} \; \ttt{a}
\; \text{(3)} \; \ttt{b}
\; \text{(4)} \; \ttt{b}
\; \text{(5)} \; \ttt{baa}$ rather than $\text{(1)} \; \ttt{a}
\; \text{(2)} \; \ttt{ab}
\; \text{(3)} \; \ttt{b}
\; \text{(4)} \; \ttt{b}
\; \text{(5)} \; \ttt{aa}$.
Note that each decomposition group corresponds to a prefix of the template: word $w_i$ is in group $\ell$ if $w_{:i}$ matches the template prefix $t_{:\ell}$.

\paragraph{Matching templates}
Our construction uses $L$ Transformer layers, where $L$ is the maximum number of states in any template.
At each layer $\ell$, we calculate whether the input matches the template prefix $t_{:\ell}$ for each template $t$ and at each position $i$.
If $t_{\ell}$ is the wildcard character \ttt{0}, then $w_{:i}$ matches $t_{:\ell}$ if $t_{:\ell - 1}$ has been matched at any position $j < i$.
If $t_{\ell}$ is a vocabulary item $w$, then $w_{:i}$ matches $t_{\ell}$ if $w_i = w$ and $w_{:i-1}$ matches $t_{:\ell - 1}$ (or, if $t_{\ell - 1}$ is \ttt{0}, if $w_{:i}$ matches $t_{:\ell - 1}$, to account for the possibility that \ttt{0} matches zero words).
We check these conditions using two attention heads per layer: %

\begin{figure}[h]
\centering
\begin{subfigure}[t]{0.49\textwidth}
\begin{minted}[fontsize=\scriptsize]{python}
def frac_prev(values, segment_ids, segment_pos):
  return aggregate(
    (select(segment_ids, segment_ids, eq) & 
     select(segment_pos, segment_pos, not_eq)),
    values)
\end{minted}
\end{subfigure}
\hfill
\begin{subfigure}[t]{0.49\textwidth}
\begin{minted}[fontsize=\scriptsize]{python}
def select_prev(values, segment_ids, segment_pos):
  return aggregate(
    (select(segment_ids, segment_ids, eq) &
     select(segment_pos, segment_pos, is_prev)),
    values)
\end{minted}
\end{subfigure}

\addtocounter{figure}{-1}

\end{figure}

These attention heads restrict attention to the most recent utterance by taking the logical AND between two selectors; see \citet[Appendix~F]{lindner2023tracr} for a discussion of mechanisms for combining selectors.
Note that each layer uses two attention heads, with each attention head calculating \ttt{frac\_prev} or \ttt{select\_prev} for all templates in parallel.

\paragraph{Templates as finite-state automata}
While our construction is presented in terms of ELIZA templates, we note that the ELIZA template language defines a subset of star-free regular languages.
As a result, we can formulate this construction as an approach to simulating a finite-state automaton, building on the constructions of~\citet{liu2023transformers} and~\citet{yang2024masked}.
In particular, consider again our example template $t = \ttt{\^{}a0bb0\$}$.
We can recognize this template by simulating the finite-state automaton illustrated in Figure~\ref{fig:template_fsa}.
\begin{figure*}
\begin{center}
\begin{tikzpicture}[shorten >=1pt,node distance=2cm,on grid,auto,initial text=] 
   \node[state] (q_0) {\ttt{1}}; 
   \node[state] (q_1) [right=of q_0] {\ttt{2}}; 
   \node[state] (q_2) [right=of q_1] {\ttt{3}}; 
   \node[state] (q_3) [right=of q_2] {\ttt{4}}; 
   \node[state] (q_4) [right=of q_3] {\ttt{5}}; 
   \node[state] (q_5) [right=of q_4] {\ttt{6}}; 
   \node[state,accepting] (q_6) [right=of q_5] {\ttt{7}}; 
    \path[->] 
    (q_0) edge  node  {\ttt{a}} (q_1)
    (q_1) edge  node  {\ttt{a}} (q_2)
          edge [bend left] node [above] {\ttt{b}} (q_3)
    (q_2) edge node {\ttt{b}} (q_3)
          edge [loop below] node {\ttt{a}} (q_3)
    (q_3) edge node {\ttt{b}} (q_4)
          edge [bend left] node [below] {\ttt{a}} (q_2)
    (q_4) edge node {\ttt{a,b}} (q_5)
          edge [bend left] node [above] {\ttt{\$}} (q_6)
    (q_5) edge node {\ttt{\$}} (q_6)
          edge [loop below] node {\ttt{a,b}} ();
\end{tikzpicture}
\end{center}
\caption{A finite-state automaton corresponding to the ELIZA template $t = \ttt{\^{}a0bb0\$}$. See App.~\ref{app:template_matching}.}
\label{fig:template_fsa}
\end{figure*}
Each state in the automaton corresponds to a prefix of the template: if the automaton is in state $\ell$ after processing words $w_1, \ldots, w_i$, then the sequence $w_{:i}$ matches the template prefix $t_{:\ell}$.
Given a template $t_1, \ldots, t_L$, we will therefore refer to the states of the corresponding automaton using the template prefixes $t_{:1}, \ldots, t_{:L}$.
Note that some special handling is required because the automaton states are assigned from left to right with no ability to look ahead in the input.
For example, consider the template \ttt{0ab} and input \ttt{bacaab}, which should be decomposed as (1) \ttt{baca} (2) \ttt{a} (3) \ttt{b}.
Without looking ahead in the input, we have no way of knowing that the first two \ttt{a} tokens belong in group 1 rather than 2.
Our template matching procedure would assign this sequence the states \ttt{121223}.
A similar issue arises if we have a template such as \ttt{01ab}, which should decompose input \ttt{bacaab} as (1) \ttt{bac} (2) \ttt{a} (3) \ttt{a} (4) \ttt{b}.
These issues can be addressed by taking some additional care in the generation stage, discussed in more detail below (App.~\ref{app:generation}).

\paragraph{Comparison to existing constructions}
Our construction differs in some ways from prior work for simulating finite state automata with Transformers.
In particular, the construction of~\citet{yang2024masked} uses hard (one-hot) attention to recognize star-free regular expressions.
Our construction uses a \ttt{frac\_prev} attention head, which attends uniformly to all positions in the sequence; this allows us to match multiple templates using one attention head.
While the number of attention heads is constant with respect to the number of templates, the embedding dimension increases linearly with the number of templates, in order to encode the automaton state for each template in parallel.

\paragraph{Reducing the number of layers}
For ease of presentation, we described a template matching construction that uses one Transformer layer for each symbol in the template.
Here, we describe two modifications that reduce the number of layers to the number of wildcard symbols in the template.

\ti{Combining wildcards:}
First, we can use one layer to match both a wildcard symbol and the symbol that immediately follows.
For example, consider the template \ttt{a0b0} and input \ttt{accbabc}, which we aim to decompose as (1) \ttt{a} (2) \ttt{cc} (3) \ttt{b} (4) \ttt{abc}.
The computations for this example are illustrated in Table~\ref{tab:combining_wildcards}.
\begin{table*}
\begin{center}
{\small
    \begin{tabular}{l c c c c c c c}
        Input & \tta & \ttc & \ttc & \ttb & \tta & \ttb & \ttc \\
        \midrule
        Attention 1 & \ttt{a} & \ttt{a0} & \ttt{a0} & \ttt{a0} & \ttt{a0} & \ttt{a0} & \ttt{a0} \\
        MLP 1 & - & - & - & \ttt{a0b} & - & \ttt{a0b} & - \\
        Attention 2 & - & - & - & - & \ttt{a0b0} & \ttt{a0b0} & \ttt{a0b0} \\
        \midrule
        Output & \ttt{1} & \ttt{2} & \ttt{2} & \ttt{3} & \ttt{4} & \ttt{4} & \ttt{4}
    \end{tabular}
}
\end{center}
\caption{
Illustration of the computations involved in template matching construction that uses one Transformer layer for each wildcard symbol in the template. 
Here, each entry in the table illustrates a value calculated at that layer, corresponding to a template prefix that has been matched at that point.
For example, the first-layer MLP identifies that the prefix \ttt{a0b} has been matched at two positions.
We distinguish between the first and second matches of this prefix by assigning each position to the longest prefix that matches at that point.
See App.~\ref{app:template_matching}.
}
\label{tab:combining_wildcards}
\end{table*}

\ti{Handling $n$-gram literals:}
The second modification pertains to $n$-gram literals in the template.
For example, consider the template \ttt{a0bc0}.
As presented above, our construction uses one layer to match the prefix \ttt{a0b} and another to match the prefix \ttt{a0bc}.
Instead, we can combine these operations into a single layer by using two attention heads.
At position $i$, one attention head checks whether the previous word $w_{i-1}$ is \ttt{b}.
The second attention head checks whether the prefix \ttt{a0} has been matched anywhere to the left of $w_{i-1}$, attending to all tokens at positions less than $i-1$.
We can use this approach for any $n$-gram up to some maximum $n$, defined by the number of attention heads per layer.

\subsection{Generating a Transformation}
\label{app:generation}

\begin{figure*}[t]
\begin{minted}[fontsize=\scriptsize]{python}
def get_reassembly_action(group_count, template, rule, step):
  # For each template t, group_count[t][l]  is the number of input tokens with group t[l]
  counts = group_count[template]

  # The position in the input sequence at the start of each group
  group_start_positions = np.concatenate([np.array([0]), np.cumsum(counts[:-1])])

  # The number of tokens in each part of the reassembly rule
  rule_part_sizes = np.array([counts[int(r)] if r.isnumeric() else 1 for r in rule])

  # The length the output will be after applying each part of the reassembly rule
  rule_part_end_positions = np.cumsum(rule_part_sizes)

  # Return to the user if we're done generating.
  if step == rule_part_sizes.sum():
    return "u:"

  # Which part of the rule are we in?
  i = np.argmax(rule_part_end_positions > step)
  r = rule[i]

  # Return the position of the token to copy: 
  if r.isnumeric():
    num_already_copied = step - (rule_part_end_positions[i - 1] if i > 0 else 0)
    target_position = int(group_start_positions[int(r)] + num_already_copied + 1)
    return "copy", target_position
  
  # Return a constant token to output.
  return "print", r
\end{minted}
\caption{
Code for generating an output token at step \ttt{i} given a user input \ttt{x}, the corresponding sequence of automaton states, and a reassembly rule.
}
\label{fig:generate_token}
\end{figure*}

Now we assume that we have identified a matching template and that the embedding for each input token identifies the decomposition group to which that token belongs.
The next step is now to apply the chosen \ti{reassembly rule} to the input to generate a response.

\paragraph{Reassembly rules}
Given a template $t_1, \ldots, t_L$ and vocabulary $\gV$, a reassembly rule is a sequence $r = r_1, \ldots, r_M$, where each $r_i$ is either a word $w \in \gV$ or an integer $n \in [M]$ such that $t_n \in \{\ttt{0}, \ttt{1}\}$.
Given an input $w_1, \ldots, w_N$, let $s_1, \ldots, s_N \in [L]$ denote the lengths of the longest matching template prefix at each position---that is, $t_{:s_i}$ is the longest prefix matching $w_{:i}$.
We refer to each $s_i$ as a \ti{decomposition group}.
For each $r_i$, if $r_i \in \gV$, the model outputs $r_i$.
If $r_i \in [L]$, the model outputs the subsequence of $w$ such that, for each $w_j$, $s_j = r_i$.
For example, consider the template $t = \ttt{a0bb0}$ and example input \ttt{aaabbab}, with automaton states \ttt{1223455}.
The reassembly rule $r = \ttt{c2d5}$ would generate the response \ttt{caadab}.
We can divide this process into two stages.
First, at each step, we need to determine the reassembly state---that is, which symbol of the reassembly rule are we currently processing.
In Fig.~\ref{fig:generate_token}, we illustrate how we can determine the state as a function of the number of tokens that have been generated so far and the number of tokens in each decomposition group.
Second, if the next token should be copied from the input, we need to identify the exact token in the input that should be copied.
We present two mechanisms for copying, one using content-based attention and one using position-based attention.

\paragraph{Option 1: Content-based attention (induction head)}
The first possible approach uses content-based attention, akin to an $n$-gram level induction head~\citep{olsson2022context,akyurek2024context}.
First, at each input position $j$, the key embedding encodes the decomposition group to which the token belongs as well as the identity of the previous $n$ tokens, where $n$ is the maximum context window.
Second, at each output position $i$, the query embedding encodes the decomposition group $s_i$ from which we should copy at this step, as well as the identity of the current token and any previous output tokens associated with this decomposition group.
An attention head can then attend to the earliest input position $j$ such that $s_j = s_i$ and, for all $k$ from 0 to $n$, if $s_{i-k} = s_i$ then $w_{j-k-1} = w_{i-k}$.
Note that we must specify a maximum context window, $n$, which is constrained by the embedding size.
If $n$ is less than the length of a decomposition group, this mechanism can fail if the same $n$-gram appears more than once in the decomposition group, as noted by~\citet{zhou2023algorithms}.

For example, consider the template $t = \ttt{a0b0}$ and reassembly rule $r = \ttt{h2}$. For an input $\ttt{acdecdfbg}$ that matches this template, the output under the reassembly rule is given by $\ttt{hcdecdf}$. If the model uses a $2$-gram induction head, the behavior of the model for the same input is given in  Tab.~\ref{tab:induction_head_reassembly}

\begin{table*}[!htbp]
\centering
{\small
 \begin{tabular}{l c c c c c c c c c c c c c c c c }
     Input & \tta & \ttc & \ttd & \tte & \ttc & \ttd & \ttf & \ttb & \ttg & \ttt{E} & \tth & \ttc & \ttd & \tte & \ttc & \ttd \\
     \midrule
     Previous $2$-gram & 00 & 0\tta & \tta \ttc & \ttc \ttd & \ttd \tte & \tte \ttc & \ttc \ttd & \ttd \ttf & \ttf \ttb  & \\
     Decomposition group & 1 & 2 & 2 & 2 & 2 & 2 & 2 & 3 & 4 & \\
     Reassembly state  &  &  &  &  &  &  &  &  &  & \tth & \ttt{2} & \ttt{2} & \ttt{2} & \ttt{2} & \ttt{2} & \ttt{2} \\
     Current $2$-gram  &  &  &  &  &  &  &  &  &  & & 00 & 0\ttc & \ttc \ttd & \ttd \tte & \tte \ttc & \ttc \ttd \\
     \midrule
     Output &  &  &  &  &  &  &  &  &  & \tth & \ttc & \ttd & \tte & \ttc & \ttd & \tte (\textcolor{red}{$\times$}) \\
 \end{tabular}%
}
\caption{
Behavior of a model that uses a $2$-gram induction head given input $\ttt{acdecdfbg}$, template $t = \ttt{a0b0}$, and reassembly rule $r = \ttt{h2}$.
Here, the model needs to output the literal token \ttt{h} and then copy the contents of the second decomposition group.
At each copying step, the 2-gram induction head attends to the position with decomposition group 2 such that the \ti{Previous 2-gram} is the longest match for the \ti{Current 2-gram}, attending to the earliest matching position in the case of ties.
For example, after generating \ttt{E h c}, the \ti{Current 2-gram} is \ttt{0c} (the previous token, \tth, is not part of this copying group, so is replaced with a \ttt{0}, which acts as a wildcard); the earliest position with the longest matching prefix is \ttt{a c d}, and the model outputs \ttt{d}.
This rule leads to an error if the same 2-gram appears more than once in the copying segment: after generating \ttt{E h} \underline{\ttt{c d}}, the model correctly outputs \ttt{e}, but after generating \ttt{E h c d e} \underline{\ttt{c d}}, the model cannot disambiguate the two occurrences of \ttt{c d} in the input and so mistakenly outputs \ttt{e}.
}
\label{tab:induction_head_reassembly}
\end{table*}

\paragraph{Option 2: Position-based attention}
Our second possible approach uses position-based attention and is described in Fig.~\ref{fig:generate_token}.
Specifically, we can use an attention head to count the number of tokens in each decomposition group, as well as the position in the input sequence at which that decomposition group begins.
A feedforward layer can then calculate the position of the input token that should be copied at a given generation step.
As discussed by~\citet{zhou2023algorithms}, this form of position arithmetic might be more difficult for the model to learn.
However, if this mechanism is learned correctly, we predict that it might generalize better than content-based attention in settings where the same $n$-gram appears multiple times in the sequence.
The behavior of the model for an input is outlined in Tab.~\ref{tab:induction_head_reassembly}.

\begin{table*}[!htbp]
\centering
{\small
 \begin{tabular}{l c c c c c c c c c c c c c c c c }
     Input & \tta & \ttc & \ttd & \tte & \ttc & \ttd & \ttf & \ttb & \ttg & \ttt{E} & \tth & \ttc & \ttd & \tte & \ttc & \ttd \\
     \midrule
     Position & 1 & 2 & 3 & 4 & 5 & 6 & 7 & 8 & 9 \\
     Decomposition group & 1 & 2 & 2 & 2 & 2 & 2 & 2 & 3 & 4 & \\
     Reassembly state  &  &  &  &  &  &  &  &  &  & \tth & \ttt{2} & \ttt{2} & \ttt{2} & \ttt{2} & \ttt{2} & \ttt{2} \\
     Position to copy &  &  &  &  &  &  &  &  &  & & 2 & 3 & 4 & 5 & 6 & 7\\
     \midrule
     Output &  &  &  &  &  &  &  &  &  & \tth & \ttc & \ttd & \tte & \ttc & \ttd & \ttf \\
 \end{tabular}%
}
\caption{
Behavior of a model that uses position-based attention  given the input $\ttt{acdecdfbg}$, template $t = \ttt{a0b0}$, and reassembly rule $r = \ttt{h2}$.
The position-based copying mechanism uses an attention head to count the number of tokens in each copy group and an MLP to calculate the target position based on current step and number of tokens per group.
Finally, an attention is used to copy the token from the target position.
}
\label{tab:position_reassembly}
\end{table*}

\begin{figure*}[t]
\centering
\centering
\includegraphics[width=\textwidth]{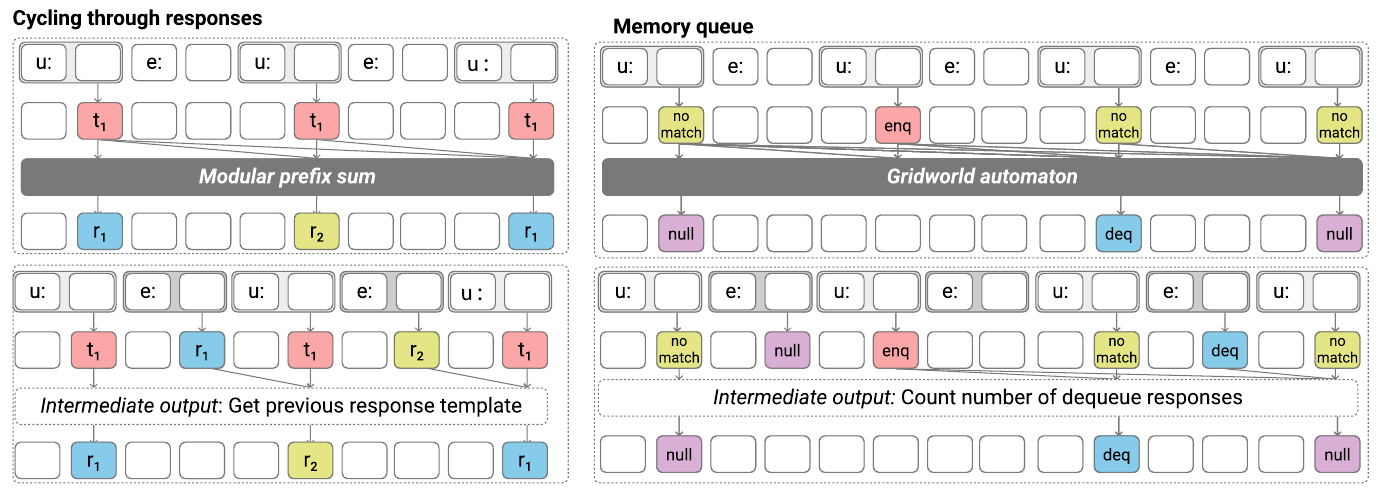}
\caption{
ELIZA includes two components that make use of the long-term conversation history: cycling through response templates (\ti{left}), and the memory queue (\ti{right}).
We identify two mechanisms for these components.
\ti{Top:} First, after parsing the user's input, we can use existing automaton constructions~\citep{liu2023transformers} as black box components to simulate the relevant data structures.
\ti{Bottom:} Alternatively, we can re-use the template matching mechanism to also parse intermediate ELIZA outputs, resulting in simpler constructions with different generalization tradeoffs.
}
\label{fig:alternatives}
\end{figure*}

\subsection{Pre-transformation Rules and an ELIZA Transformer Turing Machine}
\label{app:turing_machine}
In this section we discuss how to incorporate the special pre-transformation rule into our construction.
This rule is used by~\citet{hay2022eliza} to prove that ELIZA is Turing-complete, which will allow us to immediately derive a Turing machine construction for the ELIZA Transformer.

\paragraph{Pre-transformations with the ELIZA Transformer}
As discussed in Appendix~\ref{app:eliza_details}, a pre-transformation rule consists of a decomposition template, a transformation rule, and a reference to another keyword in the script.
If an input $w$ matches the template, ELIZA reassembles it according to the transformation rule to get a new input $w'$, and then reprocesses $w'$ according to the specified keyword.
Pre-transformation rules can trigger an arbitrary number of computational steps (for example, we can write a script corresponding to a Turing machine that never halts).
Therefore, given a Transformer with a finite number of layers, the only way to incorporate arbitrary pre-transformation rules into our construction is to enable the Transformer to perform variable computation depending on the input.
The most natural way to do this is using a Chain-of-Thought-style approach~\citep{wei2022chain}: if the input matches a pre-transformation rule, the ELIZA Transformer will output the transformed input (along with some indicator of the new state), and then reprocess the newly generated output.
This approach also follows from~\citet{merrill2024the}, who demonstrate that intermediate-decoding steps are necessary for simulating arbitrary Turing machines with decoder-only Transformers.

\paragraph{ELIZA Transformer Turing Machine}
Having incorporated pre-transformation rules into the ELIZA Transformer, we can now use the ELIZA construction from~\citet{hay2022eliza} to immediately get a new construction for simulating a Turing machine with an auto-regressive Transformer.
In this construction, each action in the Turing machine is expressed as a pre-transformation rule, and the input at each timestep encodes the tape.
Given a Turing machine (TM) that runs in $T(n)$ steps (where $n$ is the length of the input), this construction uses $T(n)^2$ generation steps: at each step, it finds the pattern that matches the most recent input, regenerates the tape according to the associated transformation rule, and then reprocesses the new version of the tape.
This resembles existing constructions, but with some differences.
For example,~\citet{wei2022statistically} give a construction that uses $T(n)$ generation steps: at each step, the model generates one new token, which encodes the state and action taken at that step.
(On the other hand,~\citet{wei2022statistically} assumes the TM uses a single-directional tape, so will take $T(n)^2$ steps to simulate a TM with a bi-directional tape running in $T(n)$ steps.) %
Note that the ELIZA construction does not use either of the long-term memory mechanisms (response cyling or the memory queue).
At each step, the model needs to attend only to the most recent version of the tape---which has a length of $T(n)$---rather than the full conversation history, which has a final length of $T(n)^2$.
The construction could therefore use sliding window attention~\citep[e.g.][]{beltagy2020longformer} to reduce the number of attention comparisons at each step.

\section{Experimental Details}
\label{app:experiments}
Here we provide more details about how we generate the data and conduct the experiments.
Code and data for reproducing the experiments will be made available after the anonymity period.

\subsection{Data Generation}
\label{app:data_generation_details}
To generate an ELIZA dataset, we first generate a set of decomposition templates and reassembly rules, and then generate conversations by generating sentences that match the different decomposition templates and applying the corresponding rules.
For all templates and sentences are drawn from a vocabulary $\gV$ consisting of the 26 lower-case English letters.
Each turn begins with a special delimiter character---\ttt{U} for user inputs and \ttt{E} for ELIZA inputs---and ends with a period, and each conversation begins with a special beginning-of-sequence token.

\paragraph{Decomposition templates}
Our distribution over decomposition templates is defined by the following parameters: the minimum and maximum number of wildcard symbols per template; and the maximum $n$-gram length, meaning the maximum number of contiguous non-wildcard symbols.
For example, the template \ttt{0a0bc0} has two wildcards and a maximum $n$-gram length of two (\ttt{bc}).
To generate a template, we first pick the number of wildcards by sampling a number $\ell$ uniformly from between the minimum and maximum, and then form a template by interleaving $\ell$ wildcard symbols with $\ell+1$ $n$-grams.
Each $n$-grams is sampled by first sampling a length $m$ uniformly from between 0 and the maximum length (for the first and last $n$-gram) or between 1 and the maximum length (for any $n$-gram between two wildcard symbols), and sampling $m$ words uniformly from $\gV$.
For our main set of experiments, we sample 31 templates with between two and four wildcards and a maximum $n$-gram length of three.
For our analysis of copying mechanisms, we sample 15 templates, each with exactly two wildcard characters and a maximum $n$-gram length of 1.
For all experiments, the final template is the null template.
The only wildcard symbol we use is \ttt{0}, corresponding to zero or more words, although ELIZA templates can also include symbols that match exactly $n$ wildcard words.

\paragraph{Reassembly rules}
Given a decomposition templates, a reassembly rule consists of a sequence of words from $\gV$ and integers indexing wildcards in the template.
We refer to these wildcards as \ti{copying segments}.
Our distribution over reassembly rules is defined by the minimum and maximum number of copying segments and the maximum $n$-gram length.
Given the set of integers corresponding to the available copying segments in the template, we generate a transformation rule by sampling up to $\ell$ of these numbers without replacement (where $\ell$ is sampled uniformly for each rule), and then form a rule by interleaving numbers with randomly sampled $n$-grams as above.
We additionally prepend each reassembly rule with a unique, constant two-word prefix.
For our main experiments, we sample up to five reassembly rules per templates, each with between one and four copying segments.
For our analysis of copying mechanisms, we sample one reassembly rule per template, each with exactly two copying segments characters.

\paragraph{Single turn}
To generate a single turn of a conversation, we sample a decomposition template and then sample a sentence that matches that template.
For each wildcard in the template, we pick a segment length $m$ uniformly from between 0 and the maximum segment length, and then sample $m$ words from the vocabulary.
For our first set of experiments, the maximum segment length is 10 and we sample the $m$ words uniformly for each segment.
In our second set of experiments, the maximum segment length is 20, and, for each segment, we first sample a unigram distribution $\vp \sim \mathrm{Dirichlet}(\alpha \mathbf{1})$, and then sampling $m$ words from $\mathrm{Categorical}(\vp)$, as described in Section~\ref{sec:mechanisms}).

\paragraph{Conversations}
For our main experiments, we generate conversations by sampling a sequence of turns until we reach the maximum input length (512 tokens).
(For our experiments with copying mechanisms, each conversation consists of a single turn.)
We take some additional considerations to ensure that the data demonstrates the cycling behavior---that is, to ensure that each template occasionally appears enough times in a conversation to cycle through all of the associated reassembly rules.
In particular, for each conversation, we sample a distribution over templates $\vp \sim \mathrm{Dirichlet}(\boldsymbol{\alpha})$, and then for each turn sample a template $t \sim \mathrm{Categorical}(\vp)$.
Here, $\boldsymbol{\alpha}$ is a 32-dimensional vector, corresponding to the 32 templates (including the null template); setting the entries of $\boldsymbol{\alpha}$ to be less than one makes it more likely that $\vp$ assigns most probability to a small number of templates.
We set the entries to be 1/32, with the exception of the memory template, which is set to 1/4 (to increase the proportion of examples that demonstrate the memory queue).
Additionally, after sampling $\vp$, we ensure that the likelihood assigned to the null template is at least half the likelihood assigned to the memory template; this is to increase the proportion of examples that contain both enqueue operations and dequeue operations (which are triggered by the null template).
For our first set of experiments, we sample 100,000 conversations for training and 20,000 for testing.
For our second set of experiments, we sample 32,000 and 16,000 conversations for training and evaluation, respectively.

\paragraph{Memory queue}
To incorporate the memory queue mechanism, we select one of the 32 templates to serve as the memory template.
This template is associated with two lists of reassembly rules: the first list is used to respond to inputs that match the template (enqueue reassembly rules), and the second list is used later in the conversation when the memory is read from the queue (dequeue reassembly rules).
In Weizenbaum's ELIZA program~\citep{weizenbaum1966eliza}, for each memory, a dequeue reassembly rule is selected at random from the list.
In our experiments, we instead use the cycling mechanism, to ensure that the behavior is deterministic.
That is, given dequeue reassembly rules $r_1, \ldots, r_M$, at the $n^{th}$ dequeue in the conversation we use the reassembly rule $r_{n \% M}$.
In our dataset, there are four dequeue reassembly rules.
We also limit the size of the queue: when sampling conversations, we ensure that the queue contains at most four memories at any time.

\subsection{Models and Training}
\label{app:models_and_training}
For all of our experiments, we train 8-layer decoder-only Transformers with 12 attention heads per layer, a hidden dimension of 768.
The models have no position embeddings but are otherwise based on the GPT-2 architecture~\citep{radford2019language} and are implemented using PyTorch~\citep{paszke2019pytorch} and HuggingFace~\citep{wolf2020transformers}.
We use the Adam optimizer~\citep{kingma2014adam} with a learning rate of 1e-4.
For multi-turn experiments, we use a batch size of 8 and train for 10 epochs.
For single-turn experiments, we use a batch size of 64 and train for 100 epochs.
For each setting, we train models with three random seeds; plots are generated with Seaborn~\citep{waskom2021seaborn} and show the 95\% confidence intervals.

\subsection{Additional Details: Mechanism Analysis}
\label{app:mechanism_analysis}

\paragraph{Cycling through responses}
Given a template $t$ with reassembly rules $r_1, \ldots, r_M$, we select conversations in which $t$ appears $n > 1$ times.
For some $i < n$, we identify the turn at which $t$ is matched for the $i^{th}$ time in the conversation, and replace the response with $r_{j}$ for some $j \neq i$.
Then we evaluate the model's response at the next occurrence of template $t$.
If the model used the modular sum, we would expect it to give the \ti{Same} response as before the intervention (responding with $r_{i+1\%M}$); if it uses the intermediate output, we would expect it to instead reply with $r_{j+1\%M}$ (\ti{Increment}).
Figure~\ref{fig:cycling_counterfactual} indicates that the model almost always increments its response, indicating that the model relies on previous responses to update the response cycle.\footnote{
The difference between \ti{Full response} and \ti{Prefix only accuracy} indicates that the model generally selects the reassembly rule predicted by the \ti{Increment} hypothesis, but does not implementing the copying step correctly, perhaps because different rules can use different decomposition groups.
}

\paragraph{Memory queue}
We conduct a similar experiment to test the memory queue mechanism.
We select conversations containing $n > 1$ two dequeue turns.
For some $i < n$, we identify the $i^{th}$ dequeue turn and replace the response with a constant string, corresponding to a null response,
and evaluate the model's response at dequeue $i+1$.
If the model used the gridworld automaton, we would expect it to give the \ti{Same} response as before, replying with memory $i+1$.
If the model relied on intermediate outputs, we would expect it to instead reply with memory $i$ (\ti{Decrement}).
Figure~\ref{fig:memory_queue_counterfactual} shows that the model almost always decrements the memory counter, indicating that it examines its own earlier responses to identify the state of the queue.

\section{Additional Results}
\label{app:additional_results}

\subsection{Errors on null inputs}
\label{app:null_inputs}

\begin{figure*}[t]
\centering
\begin{subfigure}[b]{0.57\textwidth}
    \centering
    \includegraphics[width=\textwidth]{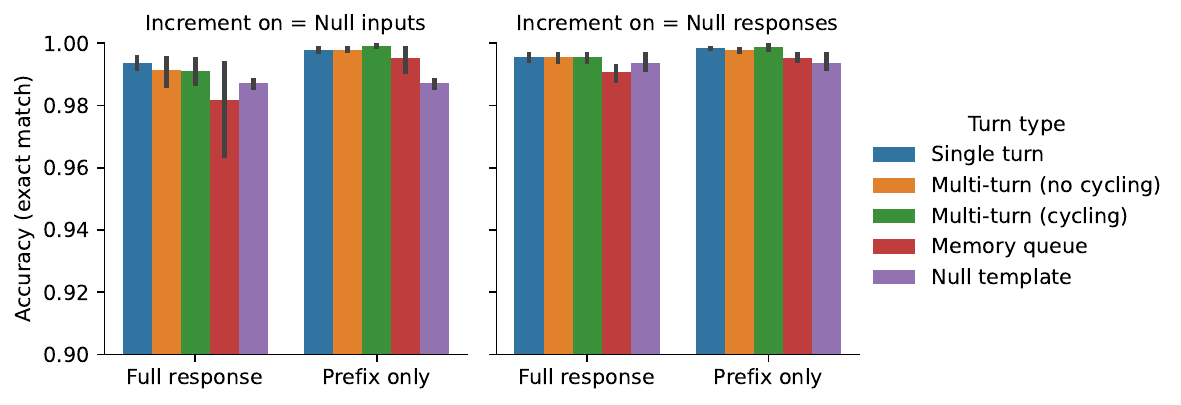}
    \caption{Accuracy (end of training).}
    \label{fig:accuracy_by_turn_type_bar_compare}
\end{subfigure}
\hfill
\begin{subfigure}[b]{0.41\textwidth}
    \centering
    \includegraphics[width=\textwidth]{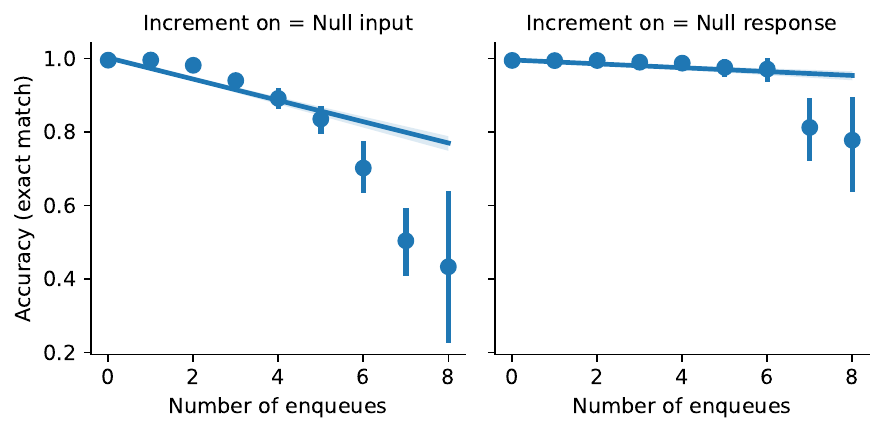}
    \caption{\ti{Null} inputs.}
    \label{fig:accuracy_null_template_compare}
\end{subfigure}
\caption{
We recreate our experiments from Sec.~\ref{sec:experiments} using a different version of the cycling mechanism for null templates.
In our original experiments, we incremented the cycle number every time the null input is matched, even if the subsequent response is to read from the memory queue.
Here, we instead increment the cycle number only when the null input is followed by a null response.
While the overall trends are similar, models trained on the second version of the data perform better overall (Fig.~\ref{fig:accuracy_by_turn_type_bar_compare}); and accuracy on null inputs does not decrease as dramatically as a function of the number of enqueues in the conversation (Fig.~\ref{fig:accuracy_null_template_compare}).
This suggests that the task is easier for models to learn when they can keep track of the cycle number using their previous responses, rather than having to count the number of null inputs.
See App~\ref{app:learning} for more details.
}
\label{fig:error_analysis_null_cycling}
\end{figure*}

In Sec.~\ref{sec:experiments}, we found that models perform worse on inputs that do not match any of the templates, in situations where the memory queue is empty.
We refer to inputs that do not match any templates as \ti{null inputs}, and say that they match the \ti{null template}.
Note that, like the other templates, the null template is associated with multiple reassembly rules, and the model should cycle through these rules when the null template is matched multiple times.
(In our experiment, there are five rules associated with the null template.)
We conjecture that the lower performance on null inputs could be related to difficulty tracking the cycle number for null templates.

In particular, there is some ambiguity in how to track the cycle number for the null template, because a null input does not always lead to a null response: if the memory queue is non-empty, the model should respond by reading from the memory queue.
In our experiments, we increment the cycle number every time the null input is matched, even if the next response reads from the memory queue.
However, we could instead increment the cycle number only when the null input is followed by a null response.
For example, consider a case where the null template is associated with three reassembly rules (``Null rule 1'', ``Null rule 2'', ``Null rule 3'').
The difference between these mechanisms is illustrated in the following conversation:

{\small
\begin{center}
    \begin{tabular}{l l l}
    \ti{User} & \ti{Cycle on input} & \ti{Cycle on response} \\
    \midrule
    U: Null. & E: Null rule 1. & E: Null rule 1. \\
    U: Memory A. & E: Enqueue. & E: Enqueue. \\
    U: Null. & E: Dequeue A. & E: Dequeue A. \\
    U: Null. & E: Null rule \tf{3}. & E: Null rule \tf{2}. \\
    \end{tabular}
\end{center}
}

We hypothesize that the first mechanism (\ti{Cycling on null inputs}) is more difficult for the model to learn; for example, the model cannot determine the cycle number by using the intermediate output mechanism described in Sec.~\ref{sec:cycling}.
To test whether this is the case, we create new conversation dataset using the same script as in our original experiments, but using the second approach to determining the cycle number for null inputs (\ti{Cycling on null responses}).
All other training details are unchanged.
The results of this experiment are plotted in Fig.~\ref{fig:error_analysis_null_cycling}.
While the error patterns are broadly similar in both cases, models trained on this second version of the data perform better overall, and do not suffer as much performance degradation as a function of the number of enqueues.
This could suggest that the task is easier for the models to learn when they can determine the cycle number as a function of previous null outputs, rather than having to count the number of null inputs.

\subsection{Copying mechanisms}
\label{app:copying_mechanisms}

\begin{figure*}[t]
\centering
\includegraphics[width=\textwidth]{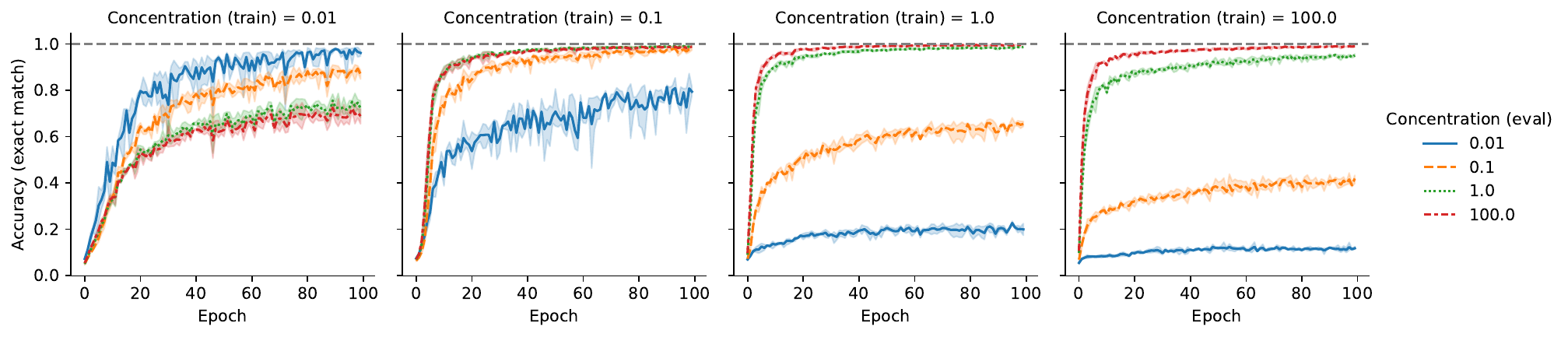}
\caption{
We train and evaluate models on datasets that vary in how likely it is for an n-gram to appear multiple times in a sequence.
These training curves correspond to the experiments discussed in \S\ref{sec:mechanisms}.
Lower values of the concentration parameter, $\alpha$, correspond to higher amounts of repetition.
For each setting, we train models with three random seeds and plot the accuracy (mean and 95\% CI) on each of the four test distributions over the course of training.
The biggest performance drop occurs when models trained with $\alpha_{\text{train}} > 0.1$ are evaluated on the setting with the most repetition ($\alpha_{\text{test}} = 0.01$); accuracy on this data also improves more slowly compared to the other settings, even when $\alpha_{\text{train}} = 0.01$ (left-most plot).
}
\label{fig:unigram_concentration_accuracy_training_curves}
\end{figure*}

\begin{figure*}[t]
\centering
\includegraphics[width=\textwidth]{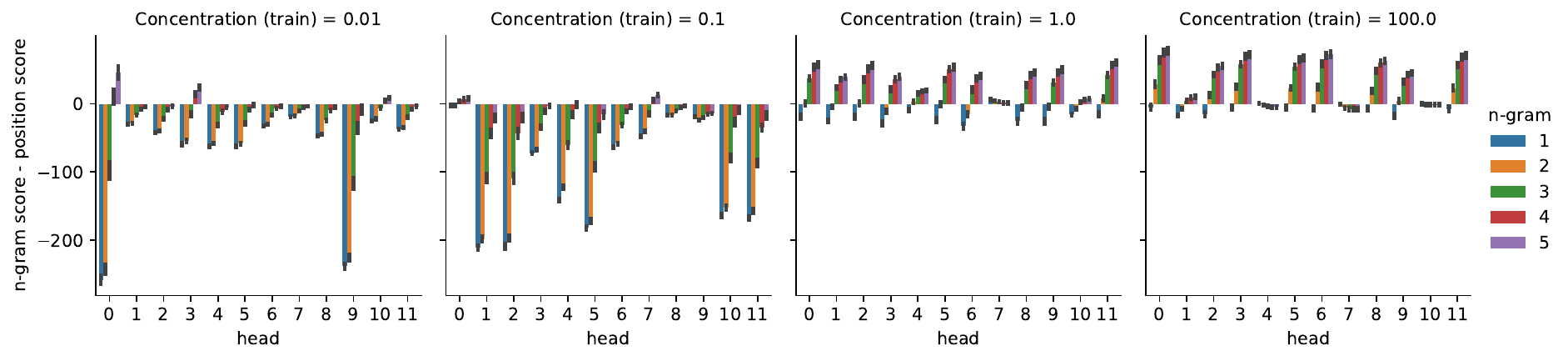}
\caption{
Which mechanism do Transformers use to copy segments of the user's input?
At each copying step, we can identify the position in the input we should read from next by counting the number of tokens in each decomposition group.
To investigate whether models use this mechanism, we compare the difference in the average attention score between queries and keys under two conditions: either the key has same $n$-gram prefix as the current output, but appears at the wrong position; or the key appears at the target position but has a different $n$-gram prefix.
In Fig.~\ref{fig:unigram_concentration_attention_difference}, we averaged this metric over all 12 attention heads in the final layer; here, we show the results for each final-layer attention head individually.
Each column corresponds to a model trained on data generated with a different concentration parameter $\alpha$, with lower values corresponding to sentences that are more likely to repeat the same $n$-grams multiple times.
For each model, the majority of attention heads show broadly similar patterns, suggesting that similar mechanisms are implemented redundantly by multiple heads.
}
\label{fig:unigram_concentration_attention_difference_by_head}
\end{figure*}

\begin{figure*}[t]
\centering
   \includegraphics[width=\textwidth]{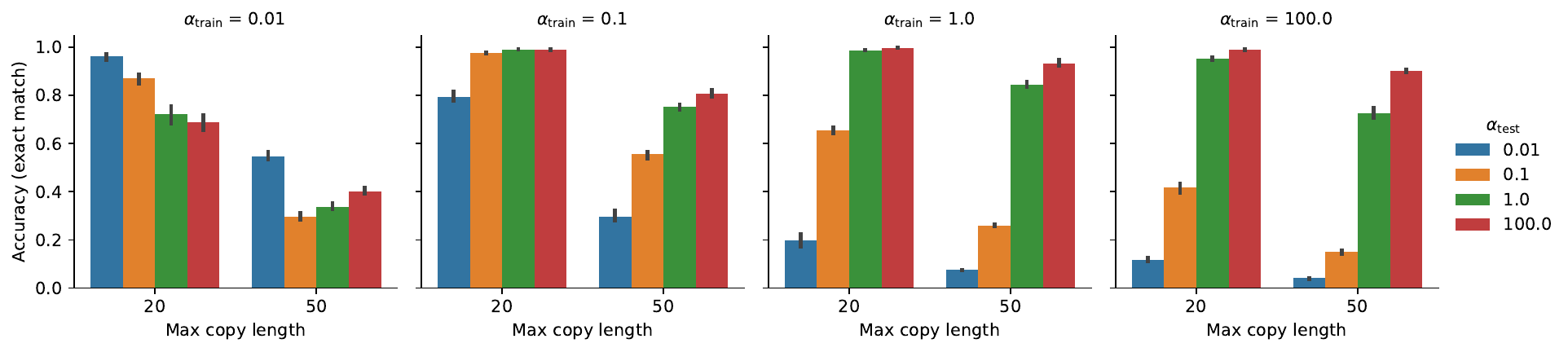}
\caption{
Models are trained on data where each copying segment has a maximum length of 20, and evaluated on data where segments can have length up to 50.
Models trained on less repetitive data ($\alpha_{\text{train}} \geq 1$) generalize worse to data with more $n$-gram repetition, but they generalize better to longer sequences.
}
\label{fig:length}
\end{figure*}

In Fig.~\ref{fig:unigram_concentration_accuracy_training_curves}, we plot the training curves corresponding to the experiments described in \S\ref{sec:mechanisms}.
Models generalize the worst to data with the highest degree of internal repetition ($\alpha_{\text{test}}=0.01$); this data also takes models longer to learn.
This agrees with the findings of~\citet{zhou2023algorithms} and could suggest that induction-head style mechanisms are easier for Transformers to learn compared to mechanisms that rely on position arithmetic.

In Fig.~\ref{fig:unigram_concentration_attention_difference_by_head}, we recreate the results from Fig.~\ref{fig:unigram_concentration_attention_difference}, but plotting the results separately for each final-layer attention head.
As discussed in \S\ref{sec:mechanisms}, in this plot, positive values indicate that the attention head has a preference for attending on the basis of position rather than content, and negative values indicate a preference for attending based on content (i.e., to tokens that have the same $n$-gram prefix as the current token), rather than position.
Interestingly, within each model, the majority of attention heads show broadly similar patterns, perhaps indicating that the models encode the same mechanism redundantly across multiple heads.
This result echoes the findings of~\citet{singh2024needs}, who find that models learn multiple parallel induction heads.
Fig.~\ref{fig:unigram_concentration_attention_difference_by_head} also suggests that none of the heads cleanly corresponds to exactly one mechanism, underscoring the challenges of aligning real-world Transformers with symbolic mechanisms.

Finally, in Fig.~\ref{fig:length}, we measure how well models generalize to data with longer copying segments.
Models trained on data with less $n$-gram repetition ($\alpha_{\text{train}} \geq 1$) generalize better to longer sequences.
This would be consistent with the claim that these models rely more on content-based attention.
As discussed in \S\ref{sec:template_matching}, we would expect content-based attention to generalize poorly to data with more $n$-gram repetition, while we would expect position-based attention to generalize poorly to data with longer copying segments.

\end{document}